\documentclass[english]{article}

\usepackage[a4paper]{geometry}
\usepackage[T1]{fontenc}
\usepackage[latin1]{inputenc}
\usepackage{amsmath,amssymb}
\usepackage{graphicx}
\usepackage{hyperref}
\usepackage{bm}
\usepackage{subfig}
\usepackage{multirow}
\usepackage{natbib}
\usepackage[noabbrev, capitalise]{cleveref}
\crefname{equation}{}{}

\newcommand{\cO}{\mathcal{O}}
\newcommand{\by}{{\mathbf y}}
\newcommand{\bx}{{\mathbf x}}
\newcommand{\bz}{{\mathbf z}}
\newcommand{\bof}{{\mathbf f}}
\def\bbR{\mathbb{R}}
\def\cX{\mathcal{X}}
\DeclareMathOperator*{\argmax}{argmax}

\title{Fast Deep Mixtures of Gaussian Process Experts}

\author{
    Clement Etienam\thanks{NVIDIA, the work in this paper was done during employment at the University of Manchester} \and
    Kody Law\thanks{University of Manchester} \and
    Sara Wade\thanks{University of Edinburgh} \and
    Vitaly Zankin \footnotemark[2] \textsuperscript{,}\thanks{The Alan Turing Institute}
}

\begin{document}
\date{}
\maketitle

\abstract{
Mixtures of experts have become an indispensable tool for flexible
modelling in a supervised learning context, allowing not only the mean function but the entire density of the output to change with the inputs. 
Sparse Gaussian processes (GP) 
have shown promise as a leading candidate for the experts in such models, and
in this article, we propose to design the gating network for selecting the 
experts from such mixtures of sparse GPs using a deep neural network (DNN). 
Furthermore, a fast one pass algorithm called Cluster-Classify-Regress (CCR) is leveraged to approximate the maximum a posteriori (MAP) estimator extremely quickly. This powerful combination of model and algorithm together delivers a novel method which is flexible, robust, and extremely efficient. In particular, the method is able to outperform competing methods in terms of accuracy and uncertainty quantification. The cost is competitive on low-dimensional and small data sets, but is {\em significantly lower for higher-dimensional and big data sets}. Iteratively maximizing the distribution of experts given allocations and allocations given experts does not provide significant improvement, which indicates that the algorithm achieves a good approximation to the local MAP estimator very fast. This insight can be useful also in the context of other mixture of experts models.
}

\section{Introduction}
\label{sec:intro}

Gaussian processes (GPs) are key components of many statistical and machine learning models. In a Bayesian setting, they provide a probabilistic approach to model unknown functions, which can subsequently be used to quantify uncertainty in predictions. An introduction and overview of GPs is given in \citep{williams2006}. 
In regression tasks, the GP is a popular prior for the unknown regression function, $f:x \rightarrow  y$, due to its nonparametric nature and tractability. It assumes that the function evaluated at any finite set of inputs $(x_1, \ldots, x_N)$ is Gaussian distributed with mean vector $(\mu(x_1), \ldots, \mu(x_N))$ and covariance matrix with elements $K(x_i, x_j)$, where the mean function $\mu(\cdot)$ and the positive semi-definite covariance (or kernel) function $K(\cdot, \cdot)$ represent the parameters of the GP.

While GPs are flexible and have been successfully applied to various problems, limitations exist. First, GP models suffer from a high computational burden, due to the need to invert and store large, dense covariance matrices. 
Second, typically parametric forms are specified for  $\mu(\cdot)$ and $K(\cdot, \cdot)$, which crucially determine properties of the regression function, such as spatial correlation, smoothness, and periodicity. This limits the model's ability to recover changing behavior of the function, e.g. different smoothness levels,  across the input space. 

To address the computational burden, one approach is to approximate based on multiple GP experts. In this case, the data is partitioned into groups, and within each group, a GP expert specifies the conditional model for the output $y$ given the input $x$.
Scalability is enhanced as each expert only  requires inversion of smaller matrices based on subsets of the data. The experts' predictions can then be combined through a product operation, known as product of experts \citep[PoEs,][]{trespBCM}, or a sum operation, known as mixture of experts  \citep[MoEs,][]{jacobs1991adaptive}. The PoE approach includes the Bayesian Committee Machine  \citep[BCM,][]{trespBCM}, generalized PoE \citep[gPoE,][]{cao2014}, and robust BCM  \citep[rBCM,][]{deisenroth2015}. PoEs are motivated by the fact that the product operation maintains Gaussianity and are specifically designed for faster inference of stationary GP models. However, a thorough theoretical analysis \citep{szabo2019asymptotic} shows that although some PoE approaches can achieve good posterior contraction rates and asymptotic coverage, this is only in the unrealistic non-adaptive setting.  Instead, MoEs work well even in adaptive settings and correspond to sound statistical models that take a weighted average of the experts' predictions. The weights are defined by the gating network, which probabilistically maps the experts to regions of the input space. The recent work of \citep{trapp2020deep} combines both operations to build a sum-product network of GPs. 

Beyond approximation, the crucial advantage of MoEs is that  model flexibility is greatly enhanced. Specifically, any simplifying assumptions of the experts need only hold for each subset of the data. This allows the model to infer different behaviors, such as smoothness and variability, within different regions, and combine the multiple experts to capture non-stationary, heterogeneity, discontinuities, and multi-modality.

In this paper, our contribution is threefold. First, we construct a novel MoE model that combines the expressive power of deep neural networks (DNNs) and the probabilistic nature of GPs. While powerful, DNNs lack the probabilistic framework and sound uncertainty quantification of GPs, and there has been increased interest in recent years in combining DNNs and GPs to benefit from the advantages and overcome the limitations of each method, see e.g. \citep{huang2015scalable, wilson2016deep, iwata2017improving, daskalakis2020faster} to name a few. Specifically, we use GP experts for smooth, probabilistic reconstructions of the unknown regression function within each region, while employing DNNs for the gating network to flexibly determine the regions.  To further enhance scalability, we combine the distributed approximation of GPs through the MoE architecture with low-rank approximations using an inducing point strategy \citep{snelson2006}. This combination leads to a robust and efficient model which is able to outperform competing models.

Second, we provide a fast and accurate approximation of the proposed deep mixture of sparse GP experts, using a recently introduced method called Cluster-Classify-Regress  \citep[CCR,][]{EtienamLaw}.
 Finally, a novel connection is made between CCR and optimization algorithms 
 commonly used for MAP estimation of MoEs, which lends credibility to 
 its success as an approximation algorithm in this context. 
 More generally, the connection both provides a framework to potentially further refine the CCR solution through additional iterations of the optimization algorithm and also sheds lights on the MoE model underlying the CCR algorithm, which provides a fast approximation for other MoE architectures as well.

\section{Methodology}\label{sec:model}

For i.i.d. data $(y_1, \ldots y_N)$, mixture models are an extremely useful tool for flexible density estimation due to their ability to approximate a large class of densities and their attractive balance between smoothness and flexibility. When additional covariate information is present and the data consists of input-output pairs, $ \{(x_i,y_i)\}_{i=1}^N$, MoEs extend mixtures to achieve flexible conditional density estimation (also known as density regression), where the whole density of the output changes with the inputs. This is achieved by modelling the mixture parameters as functions of the inputs, that is, by defining the gating network, which probabilistically partitions the input space  into regions, and by specifying the experts, which characterize the relationship between $x$ and $y$ within each region. This results in flexible framework which has been employed in numerous applications; for a recent 
overview, see \citep{Gormely2019}.

Specifically, the MoE model assumes that outputs are independently generated, for $i = 1,\ldots, N$, from a mixture:
\begin{equation}\label{eq:MoE}
\begin{aligned}
y_i \mid  x_i &\sim \sum_{l=1}^L w_l(x_i ; \psi)  \text{N}\left(y_i \mid f(x_i; \theta_l),\sigma^{2 }_l\right),\\ 
\end{aligned}
\end{equation}
where $w_l(\cdot ; \psi)$ is the gating network with parameters $\psi$; 
$f(\cdot ; \theta_l)$ is the regression function for the $l^\text{th}$ expert with parameters $\theta_l$; and $L$  is the number of experts. For simplicity, we focus on the case when $y \in \mathbb{R}$ and employ a Gaussian likelihood for the experts, although this may be generalized for other data types.  The gating network  $(w_1(\cdot ; \psi), \ldots, w_L(\cdot ; \psi))$, which maps the input space $\mathcal{X} \subseteq \bbR^d$ to the $L-1$ dimensional simplex (i.e. $w_l(x; \psi)\geq 0$ and $\sum_{l=1}^L w_l(x; \psi) =1)$, reflects the relevance of each expert at any location $x \in \mathcal{X}$.

The experts form the building blocks of the mixture model, which are combined to recover general shapes for the conditional density of $y$ given $x$, that are too complicated to be captured by a single expert. To provide intuition, MoEs can be used when the population consists of (input-dependent) sub-populations, such that within each, a single expert provides a good description of the relationship between $y$ and $x$.  However, in general, the groups may not represent actual sub-populations but rather they are combined to achieve flexibility and approximate a wide class of conditional densities. 
MoEs can be augmented with a set of allocation variables $\bz =(z_1, \ldots, z_N)$ indicating group or cluster membership, 
where $z_i = l$ if the $i^\text{th}$ data point is generated from the $l^\text{th}$ expert, for $l = 1, \ldots, L$. Specifically, if 
\begin{align*}
	p(z_i = l \mid x_i) = w_l (x_i, \psi) \quad \text{and} \quad p(y_i \mid x_i, z_i) = \text{N}\left(y_i \mid f(x_i; \theta_{z_i}),\sigma^{2 }_{z_i}\right),
	\end{align*}
then \cref{eq:MoE} is recovered after marginalization of $z_i$:
\begin{align*}
 p(y_i \mid x_i) &= \sum_{l=1}^L p(y_i, z_i=l \mid x_i) = \sum_{l=1}^L p(z_i=l \mid x_i)  p(y_i \mid x_i, z_i=l) \\
		&= \sum_{l=1}^L w_l (x_i, \psi)   \text{N}\left(y_i \mid f(x_i; \theta_{l}),\sigma^{2 }_{l}\right).
\end{align*}
Thus, letting $\by=(y_1,\ldots, y_N)$ and $\bx = (x_1, \ldots, x_N)$, we can define the augmented model as
\begin{align*}
p(\by, \bz \mid  \bx) &= \prod_{i=1}^N p(y_i \mid  x_i, z_i) p(z_i \mid x_i) =  \prod_{i=1}^N w_{z_i} \text{N}(y_i \mid f(x_i; \theta_{z_i}),\sigma^{2}_{z_i}) \\
&= \prod_{l=1}^L  \prod_{i:z_i=l} w_l (x_i ; \psi) \text{N}(y_i \mid f(x_i; \theta_{l}),\sigma^{2}_l).
\end{align*}
This augmented version of the model is widely used in inference algorithms (such as expectation-maximization), as direct optimization of the likelihood in \cref{eq:MoE} is challenging, due to multi-modality and well-known identifiability issues associated with mixtures. Moreover, the introduction of the allocation variables $\bz$ permits distributed inference across experts.

\begin{figure*}[ht]
\centering
	\subfloat[True allocations]{\includegraphics[width=0.245\textwidth]{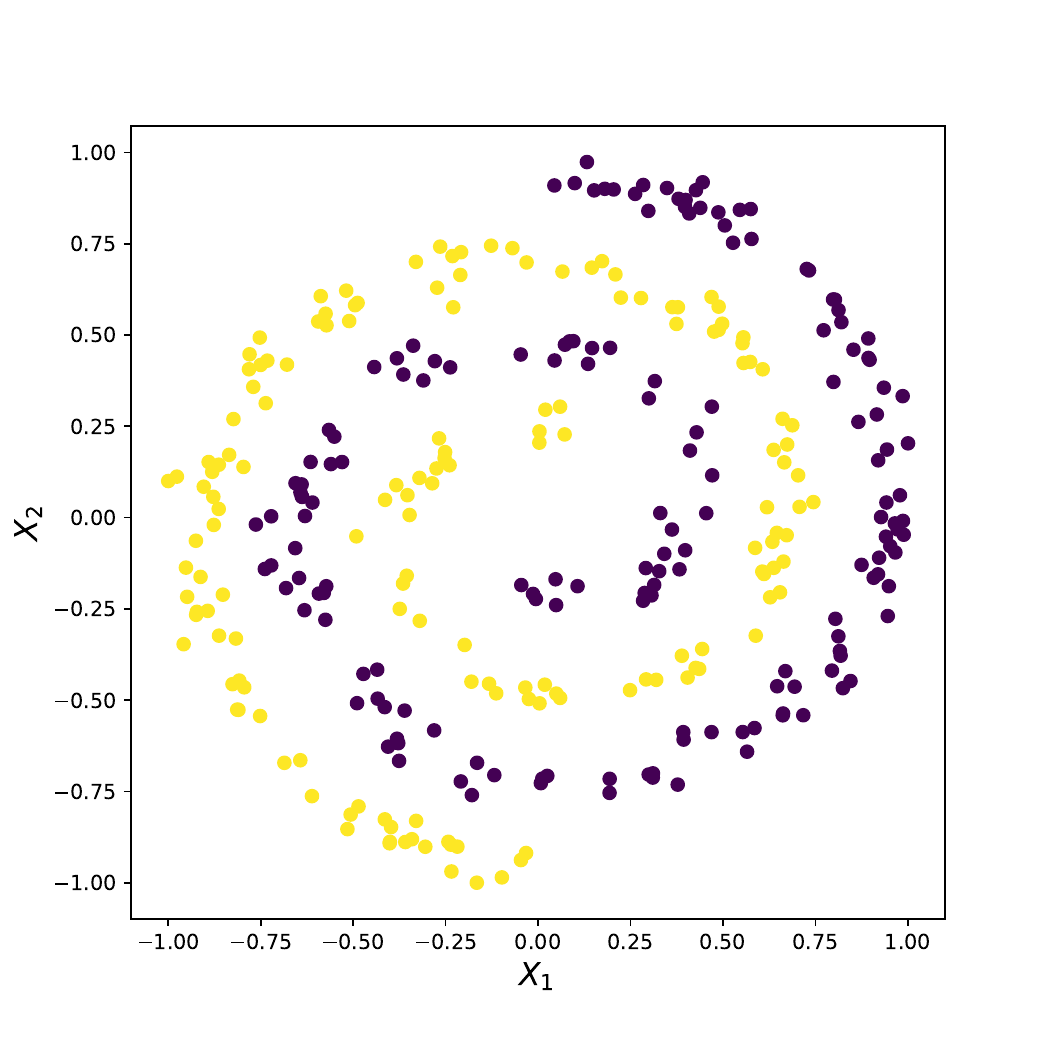}}
	\subfloat[DNN+GP allocations]{\includegraphics[width=0.245\textwidth]{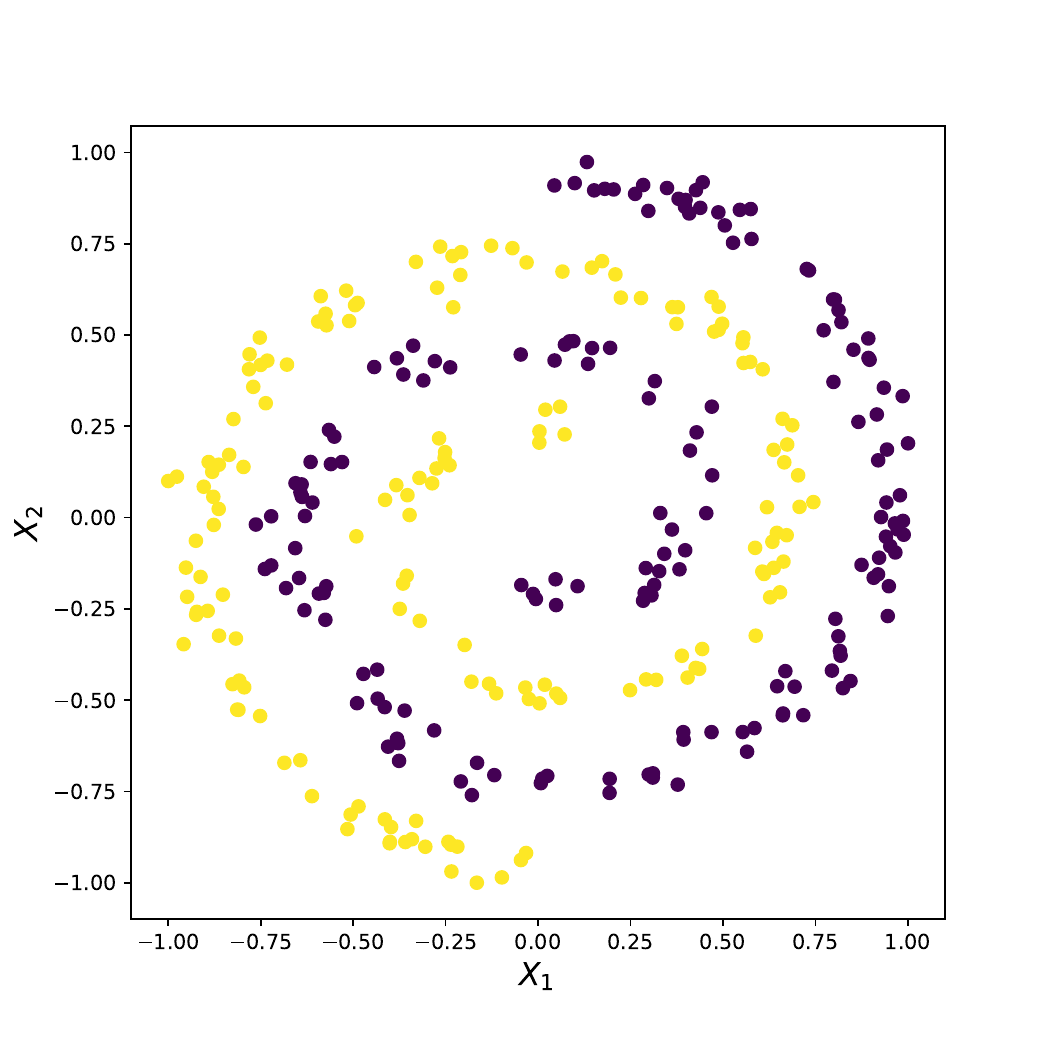}}
	\subfloat[LR+GP allocations]{\includegraphics[width=0.245\textwidth]{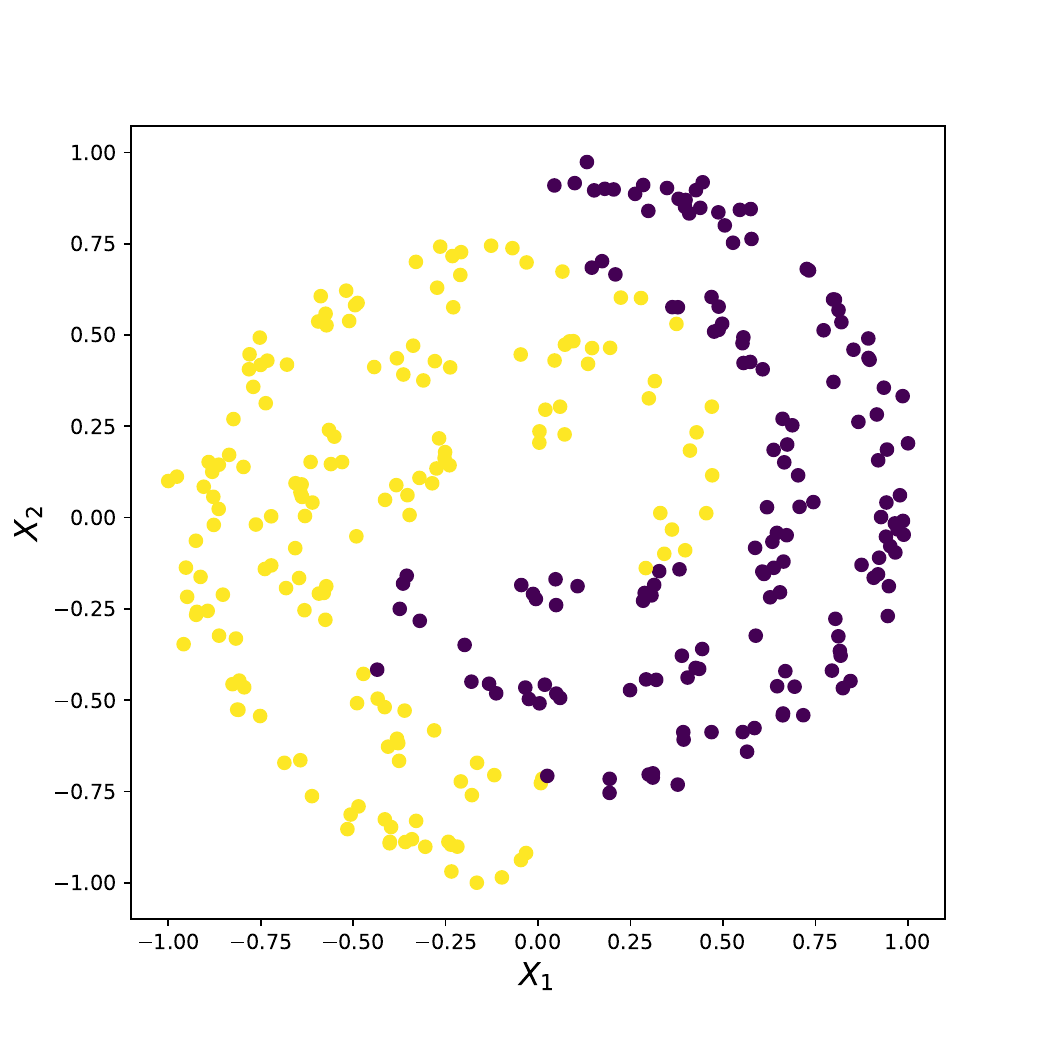}}
	\subfloat[MDN allocations]{\includegraphics[width=0.245\textwidth]{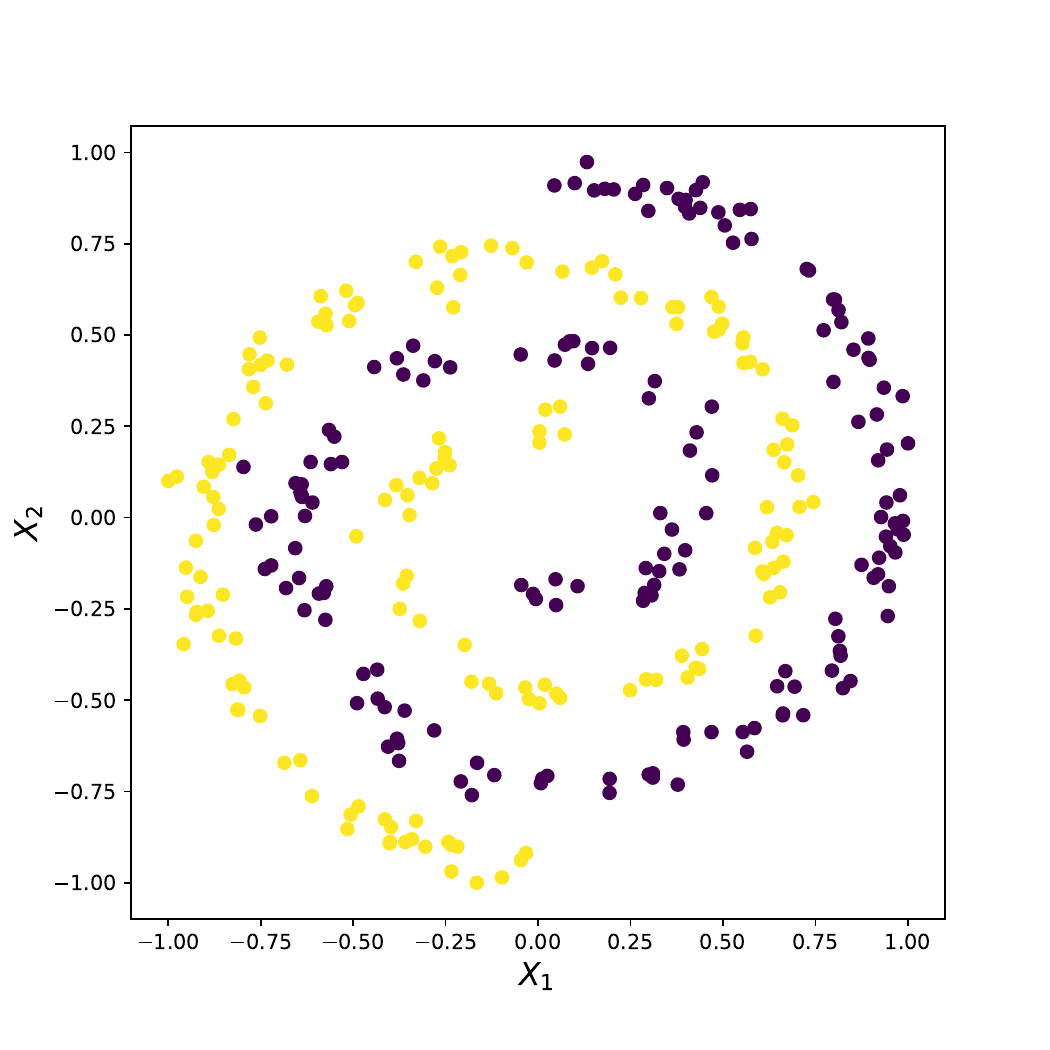}}\\
	\subfloat[True predictions]{\includegraphics[width=0.245\textwidth]{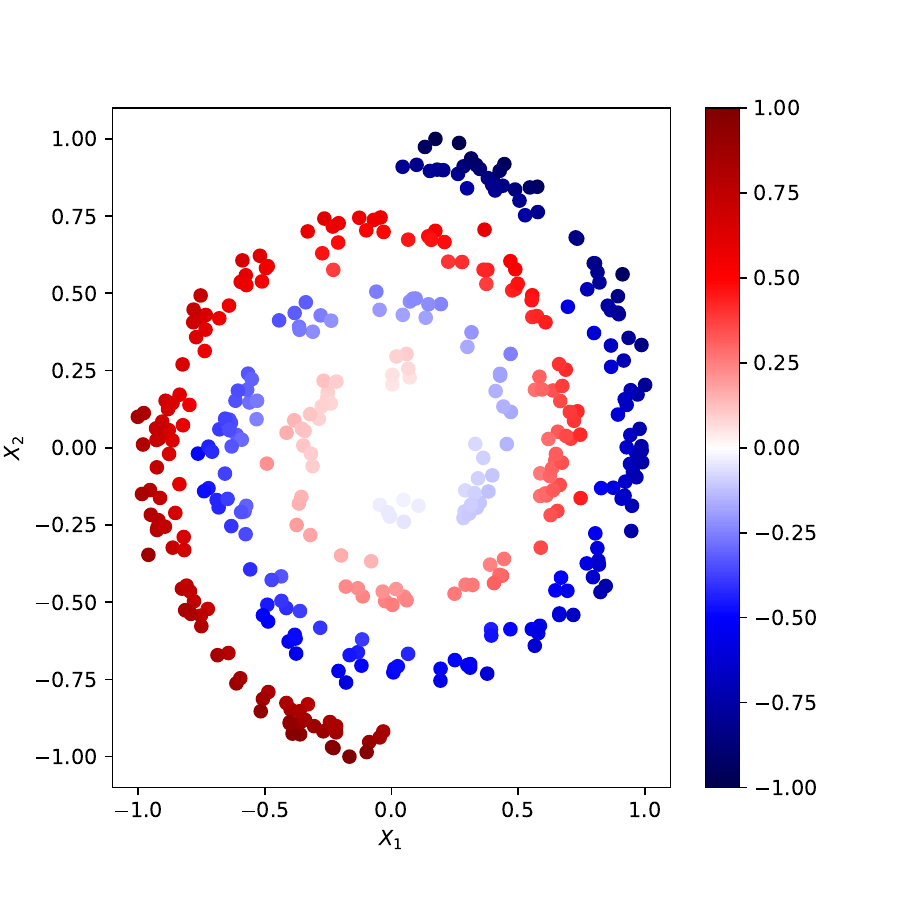}}
	\subfloat[DNN+GP predictions]{\includegraphics[width=0.245\textwidth]{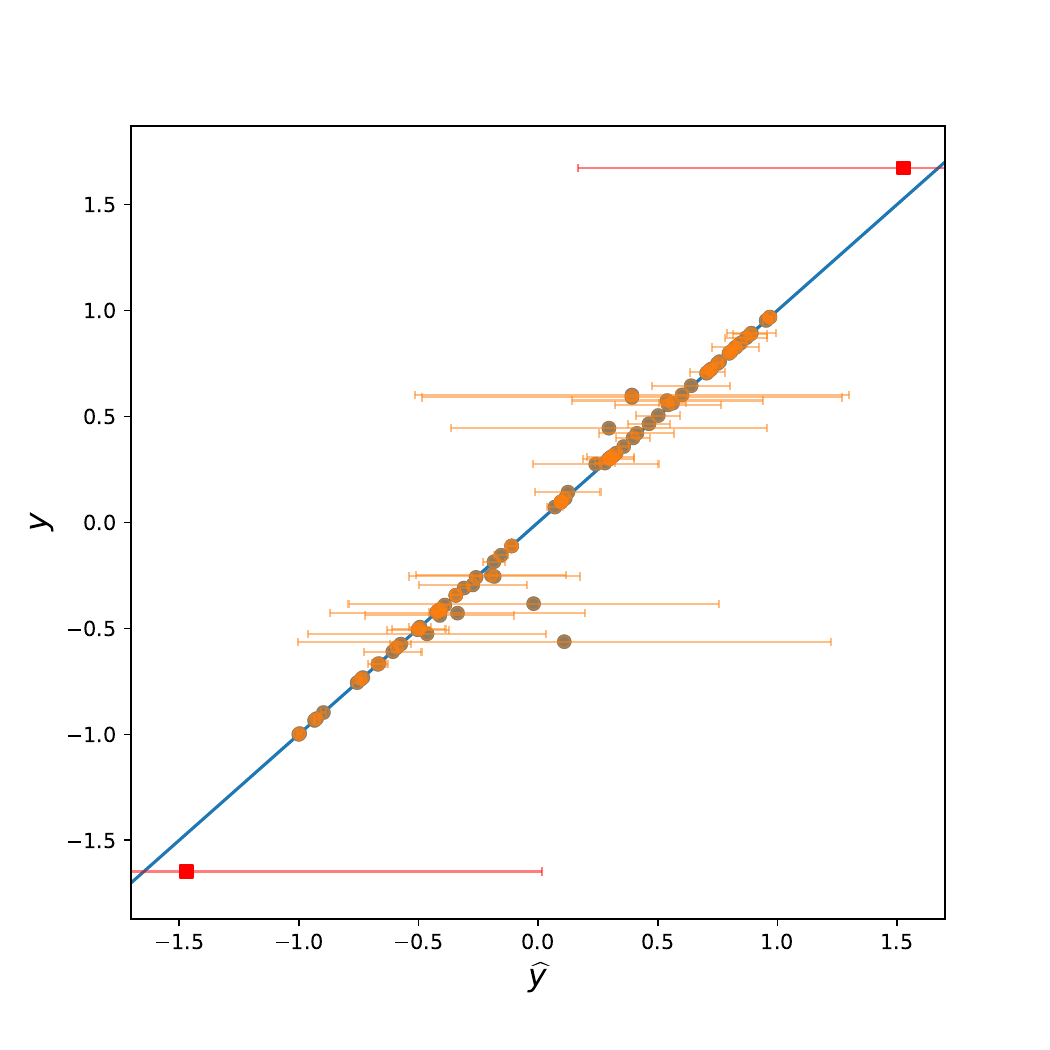}}
	\subfloat[LR+GP predictions]{\includegraphics[width=0.245\textwidth]{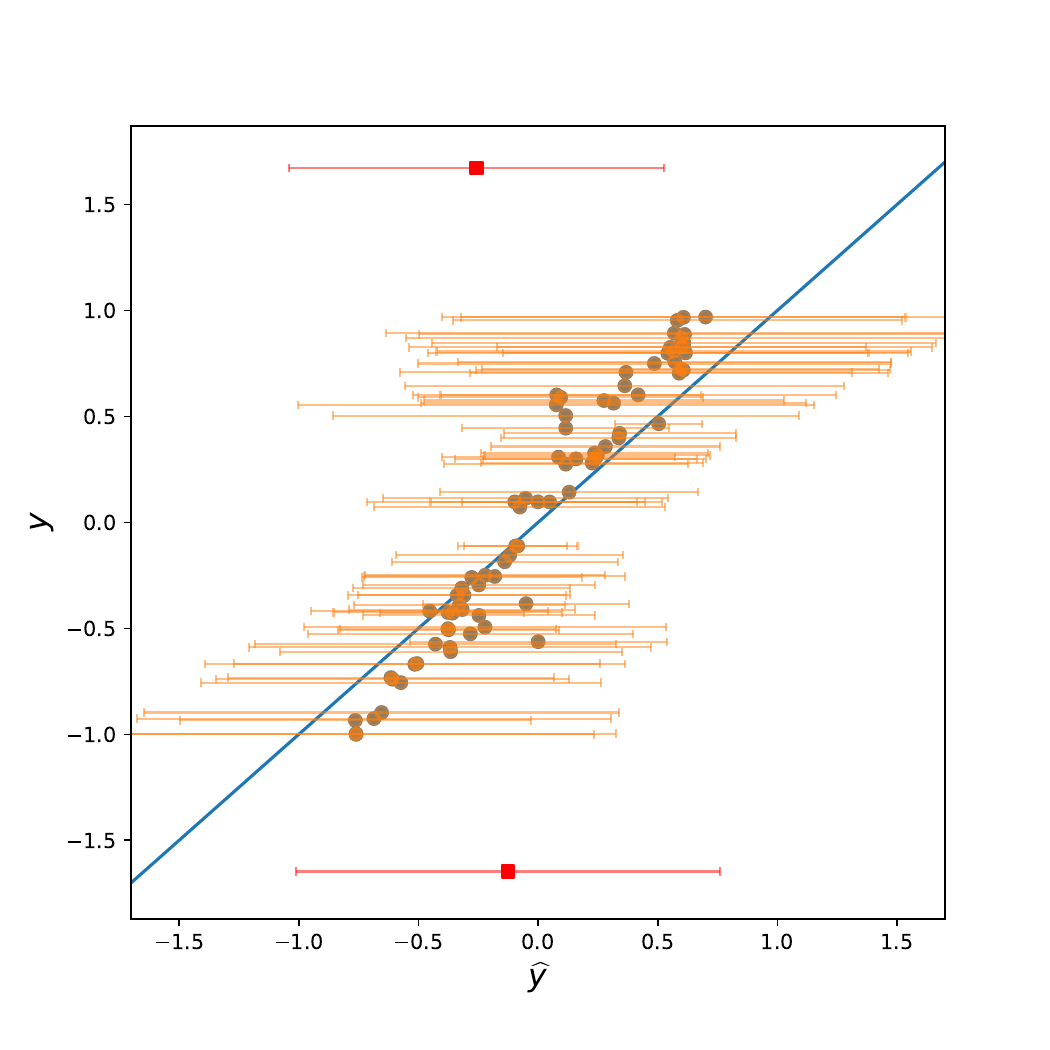}}
	\subfloat[MDN predictions]{\includegraphics[width=0.245\textwidth]{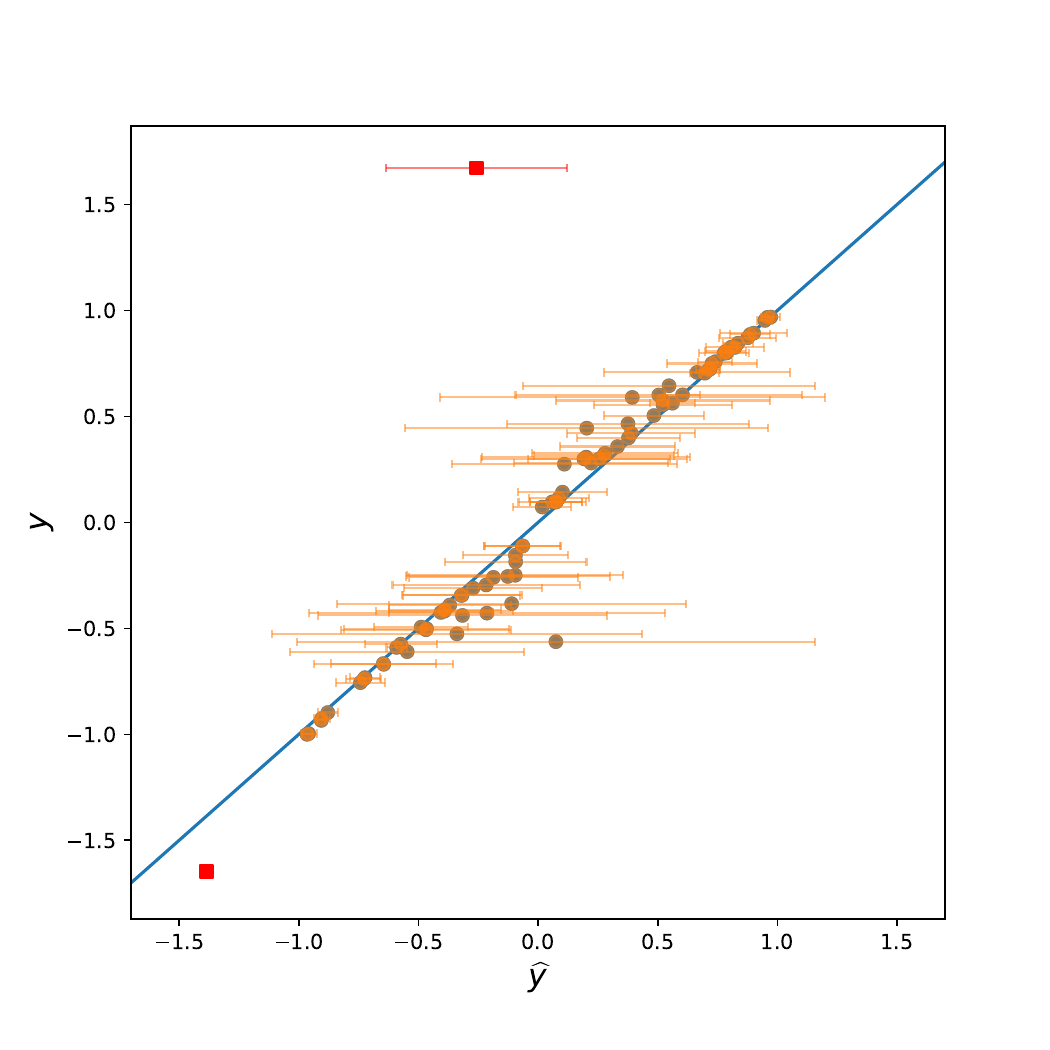}}
	\caption{Motivating toy example. Comparison of the true and estimated allocations (first row) and predictions (second row) for the proposed DNN gating network with GP experts (second column), logistic regression (LR) gating network with GP experts (third column), and DNN gating network and experts (fourth column). The DNN gating network recovers the true allocations, and combined with GP experts leads to improved accuracy and uncertainty  (details in \cref{sec:simulations}), especially for outlying test points.}
	\label{fig:toy_example}
\end{figure*}

Various formulations have been proposed in literature for both the experts and gating networks, ranging from simple linear models to flexible nonlinear approaches.  Examples include (generalized) linear or semi-linear models \citep{jordan1994,xu1995}, splines \citep{rigon2017}, neural networks \citep{bishop1994, ambrogioni2017}, GPs \citep{tresp2001,rasmussen2002infinite}, tree-based classifiers \citep{gramacy2008bayesian}, and others. 
In this work, we combine sparse GP experts with DNN gating networks to flexibly determine regions and provide a probabilistic nonparametric model of the unknown regression function. \cref{fig:toy_example} highlights the advantages of this construction: 1) flexible regions (over quadratic classifiers, e.g \citep{meeds2006, nguyen2014, zhang2019embarrassingly}) and 2) well-calibrated uncertainty (over DNN experts, e.g  \citep{bishop1994,ambrogioni2017}), through tighter credible intervals that maintain the desired coverage.

\subsection{Sparse Gaussian process experts}

Mixtures of GP experts have proven to be very successful 
\citep{tresp2001, rasmussen2002infinite, meeds2006, yuan2009, nguyen2014, gadd2019}. In particular, they overcome limitations of stationary GPs by reducing  computational complexity through distributed approximations and allowing different properties of the function for each GP expert to handle challenges, such as discontinuities, non-stationarity, non-Gaussianity and multi-modality. They allow flexible conditional density estimation, going beyond the iid Gaussian error assumption of standard GP regression models.

 In this case, one assumes a GP prior on the regression function for each expert with expert-specific hyperparameters 
$ \theta_l =( \mu_l,\phi_l)$:
 $$ f(\cdot; \theta_l) \sim \text{GP}(\mu_l, K_{\phi_l}),$$
 where $\mu_l$ is the mean function of the expert (for simplicity, it assumed to be constant) and $\phi_l$ are the parameters of the covariance function $K_{\phi_l}$, whose chosen form and hyperparameters encapsulate properties of the function for each expert. 
While it is common to use zero mean functions in  standard GP models, which is made appropriate by first subtracting the overall mean from the output, in mixtures, we must include a constant mean, as the clustering structure is unknown and the data cannot be centred for each expert. Letting  $f_{l,i} =f(x_i; \theta_{l})$ denote the $l^\text{th}$ function evaluated at $x_i$ with $\bof_l = \lbrace f_{l,i} \rbrace_{z_i = l}$, the GP prior implies a multivariate normal prior on the function $\bof_l$ evaluated at the inputs within the $l^\text{th}$ cluster:
 \begin{align*}
 	\bof_l  \sim \text{N}( \bm{\mu}_l, \mathbf{K}_{l,N_l}),
 	\end{align*}
 where  $\bm{\mu}_l $ is a vector with entries $\mu_l$; $\mathbf{K}_{l,N_l}$ represents the $N_l \times N_l$ matrix obtained by evaluating the covariance function $K_{\phi_l}$ at each pair of inputs in the $l^\text{th}$ cluster; and $N_l$ is number data points in the $l^\text{th}$ cluster. 
 With the Gaussian likelihood, the unknown function can be marginalized, and the likelihood of the data within each cluster is:
 \begin{align*}
 	p(\by_l \mid  \bx_l) &= \int \prod_{i:z_i=l} \text{N}(y_i \mid f_{l,i} ,\sigma^{2}_l) \text{N}(\bof_l \mid \bm{\mu}_l, \mathbf{K}_{l,N_l}) d\bof_l\\
 	&= \text{N}(\by_l \mid \bm{\mu}_l, \mathbf{K}_{l,N_l}+\sigma^{2}_l \mathbf{I}_{N_l} ),
\end{align*}
where $\by_l$ and $\bx_l$ contain the outputs and inputs of the $l^\text{th}$ cluster, i.e. $\by_l = \lbrace y_i \rbrace_{z_i=l}$ and $\bx_l = \lbrace x_i \rbrace_{z_i=l}$.
 
GP experts are appealing due to their flexibility, intrepretability and probabilistic nature, but they come with an increased computational cost. Specifically, given the allocation variables $\bz$,
the GP hyperparameters, which crucially determine the behavior of the unknown function, can be estimated by optimizing the log marginal likelihood:
\begin{align*}
\log(p(\by \mid  \bx, \bz)) = \sum_{l=1}^L  \log\left( \text{N}(\by_l \mid \bm{\mu}_l, \mathbf{K}_{l,N_l}+\sigma^{2}_l \mathbf{I}_{N_l} )\right).
\end{align*}
This requires inversion of $N_l \times N_l$ matrices. Compared with standard GP models, the cost is reduced from $\cO(N^3)$ to $\cO(\sum_{l=1}^L N_l^3)$, however this can still be expensive, depending on the size of clusters.

To further improve scalability, one can resort to approximate methods for GPs, including sparse GPs \citep{snelson2006, titsias2009, bui2016}, predictive processes  \citep{banerjee2008}, 
basis function approximations \citep{cressie2008}, or sparse formulations of the precision matrix \citep{lindgren2011, grigorievskiy2017, durrande2019}, 
among others (see reviews in Chp. 8 of \citep{williams2006} and \citep{heaton2019}).  
In the present work, we employ sparse GPs, which augment the model with a set of $M_l<N_l$ pseudo-inputs $\tilde{\bx}_l = (\tilde{x}_{l,1}, \ldots, \tilde{x}_{l,M_l})$ 
and pseudo-targets $\tilde{\bof}_l= (\tilde{f}_{l,1}, \ldots, \tilde{f}_{l,M_l}) $ representing the $l^\text{th}$ function evaluated at the pseudo-inputs, i.e. $ \tilde{f}_{l,m} = f(\tilde{x}_{l,m}; \theta_l)$ for $m=1, \ldots, M_l$. Then, the key assumption for scalability is that the prior on the regression function within each cluster factorizes given the pseudo-targets:
\begin{align*}
p(\bof_l, \tilde{\bof}_l \mid  \bx_l, \tilde{\bx}_l ) &= p(\bof_l \mid \tilde{\bof}_l ,  \bx_l, \tilde{\bx}_l ) p(\tilde{\bof}_l \mid  \tilde{\bx}_l )\\
&\approx \prod_{i:z_i=l}  p(f_{l,i}  \mid \tilde{\bof}_l ,  \bx_l, \tilde{\bx}_l) p(\tilde{\bof}_l \mid  \tilde{\bx}_l ).
\end{align*}
From properties of GPs, the prior of the pseudo-targets is $\tilde{\bof}_l \sim \text{N}(\bm{\mu}_l, \mathbf{K}_{l,M_l})$, and the augmented model for $\bof_l$  given the pseudo-targets is: 
\begin{align*}
	 p(\bof_l \mid  \bx_l, \tilde{\bx}_l , \tilde{\bof}_l) = \prod_{i:z_i=l} \text{N}(f_{l,i} \mid  \widehat{\mu}_{l,i}, 
 \lambda_{l,i}), 
 \end{align*}
where
\begin{align}
\widehat{\mu}_{l,i} &= 
\mu_l + (\mathbf{k}_{l,M_l,i})^T(\mathbf{K}_{l,M_l})^{-1} (\tilde{\bof}_l- \bm{\mu}_l), \label{eq:sparseGP_pred}\\
\lambda_{l,i} &=  K_{\phi_l}(x_i,x_i)- 
(\mathbf{k}_{l,M_l,i})^T(\mathbf{K}_{l,M_l})^{-1}\mathbf{k}_{l,M_l,i}, \label{eq:sparseGP_pred2}
\end{align}
with $\mathbf{K}_{l,M_l}$ denoting the $M_l \times M_l$ matrix with elements $K_{\phi_l}(\tilde{x}_{l,j},\tilde{x}_{l,h})$ and $\mathbf{k}_{l,M_l,i}$ denoting the vector of length $M_l$ with elements $K_{\phi_l}(\tilde{x}_{l,j}, x_i)$. 
This corresponds to the fully independent training conditional (FITC) approximation \citep{quinonero2005} (see \citep{bauer2016} for a further discussion of FITC). 
For the Gaussian likelihood, $\bof_l$ can be marginalized and the likelihood of the data points within each cluster also factorizes as:
\begin{align} p(\by_l \mid  \bx_l, \tilde{\bx}_l , \tilde{\bof}_l) = \prod_{i:z_i=l} \text{N}(y_i\mid  \widehat{\mu}_{l,i}, 
	\sigma^{2 }_l +\lambda_{l,i} ). \label{eq:sparse_like}
\end{align}
After marginalization of the pseudo-targets, the pseudo-inputs $\tilde{\bx}_l $ and hyperparameters $(\mu_l, \phi_l)$ can be estimated by optimizing the marginal likelihood:
\begin{equation}\label{eq:sparse_marg}
\begin{aligned}
&\log(p(\by \mid  \bx, \bz, \tilde{\bx})) = \sum_{l=1}^L  \log\left( \text{N}(\by_l \mid \bm{\mu}_l, \bm{\Sigma}_l ) \right)\, ,
\end{aligned}
\end{equation}
where $ \bm{\Sigma}_l = (\mathbf{K}_{l,M_l N_l})^T(\mathbf{K}_{l,M_l})^{-1} \mathbf{K}_{l,M_l N_l}+ \bm{\Lambda}_l +\sigma^{2}_l \mathbf{I}_{N_l}$; 
 $\mathbf{K}_{l,M_l N_l}$ is the $M_l \times N_l$ matrix with columns $\mathbf{k}_{l,M_l,i}$; and $\bm{\Lambda}_l$ is the diagonal matrix with diagonal entries $\lambda_{l,i}$. 
 This strategy allows us to reduce the complexity to $\cO(\sum_{l=1}^L N_l M_l^2)$.

\subsection{Deep neural gating networks}

The choice of gating network is crucial to flexibly determine the regions and accurately approximate the conditional densities. Indeed, various proposals exist to combine GP experts with different gating networks, and approaches to specify the gating network can be divided into generative or discriminative classifiers. Generative models specify a joint mixture model for $y$ and $x$ and typically assume a Gaussian distribution for the inputs within each class, resulting in linear or quadratic discriminant analysis, e.g.  \citep{meeds2006, chen2014precise, zhang2019embarrassingly}. To allow more flexibility in the regions, \citep{yuan2009,gadd2019} extend this by considering a mixture of Gaussians within each class. Discriminative models, on the other hand, avoid modelling the inputs and focus on the conditional density of interest by defining the gating network directly.  
Discriminative models include linear classifiers \citep{nguyen2014} and tree-based classifiers \citep{gramacy2008bayesian}, which may result in rigid assumptions on the partition structure of the input space. The latter, for example, assumes axis-aligned rectangular regions, and while more flexible partitioning approaches exist, such as Voronoi tessellations \citep{pope2019gaussian}, they come at an increased cost. Other flexible proposals include GP classifiers \citep{tresp2001} and kernel-based methods \citep{rasmussen2002infinite}, but these similarly suffer from the trade-off between flexibility and cost.

In this work, we focus on DNNs, which are known to be {\em universal for classification} \citep{szymanski2012deep}.
DNNs flexibly determine the regions, removing any rigid assumptions, without increasing the computational cost. 
Specifically, we define the gating network by a feedforward DNN with a softmax output:
$$
w_l(x ; \psi) = \frac{\exp(h_l(x ; \psi))}{\sum_{j=1}^L \exp(h_j(x ; \psi))} \, , 
$$
where $h_l$ is the $l^{\rm th}$ component of $h: \bbR^d \rightarrow \bbR^{L}$, defined 
by 
$$
h(~\cdot~ ; \psi) = 
\eta_{J}( \eta_{J-1}( \cdots \eta_{1}( ~\cdot~ ; \psi_{1}) \cdots ; \psi_{J-1}) ; \psi_{J}) \, , 
$$
with $\eta_{j}: \bbR^{d_{j-1}} \rightarrow \bbR^{d_j}$ ($d_0=d$, $d_J=L$) 
the $j^{\rm th}$ layer of a neural network 
$$\eta_{j}(~\cdot~; \psi_{j}): x \mapsto  \eta_{j}(x; \psi_{j}) = {\rm ReLU}(A_{j}x + b_{j})\, ,$$ 
where ${\rm ReLU}(x) = {\rm max} \{0, x\}$ is the element-wise rectifier; 
 $\psi_{j} = \{A_{j}, b_{j}\}$ comprises the weights $A_{j} \in \bbR^{d_{j} \times d_{j-1}}$
and biases $b_j \in \bbR^{d_j}$ for level $j = 1, \dots, J$; and $\psi = (\psi_1,\ldots, \psi_J)$ collects all the parameters of the DNN.

Deep neural gating networks have been used in literature but are typically combined with DNN experts \citep{bishop1994, ambrogioni2017}. 
The mixture density network  \citep[MDN,][]{bishop1994} uses this gating network but parametrizes both the regression function and variance of the Gaussian model in \cref{eq:MoE} by DNNs. 
This offers considerable flexibility beyond standard DNN regression, 
but significant valuable information can be gained with GP experts. 
Specifically, as the number of data points in each cluster is data-driven, DNN experts may overfit due to small cluster sizes.
In addition, DNNs are susceptible to adversarial attacks and tend to be overconfident even when predictions are incorrect \citep{szegedy2013intriguing}. Moreover, DNNs may underestimate variability, especially for test data which are quite different from the observed data \citep{nguyen2015deep} (see \cref{fig:toy_example}). 
Instead, GP experts probabilistically model the local regression function, avoiding overfitting and providing well-calibrated uncertainty quantification.

\section{Inference}

While Markov chain Monte Carlo algorithms provide full Bayesian inference for MoEs, they rely on expensive  Gibbs samplers \citep{rasmussen2002infinite, meeds2006, gadd2019}, which alternately sample the allocation variables $\bz$ given the model parameters (that is, the gating and experts parameters) and the model parameters given $\bz$.
To alleviate the cost, \citep{zhang2019embarrassingly} use importance sampling and parallelization. However, accurate estimation requires the number of important samples to be exponential in the Kullback-Leibler divergence between the proposal and posterior \citep{chatterjee2018sample}; this can be prohibitive for MoEs due to the massive dimension of the partition space. A popular alternative strategy is approximate inference, where the Gibbs sampling steps for allocation variables and the model parameters are replaced, respectively, with either an expectation or maximization step \citep{kurihara2009bayesian}. An expectation-expectation strategy corresponds to mean-field variational Bayes for MoEs \citep{yuan2009}.  While popular expectation-maximization (EM) algorithms can be used for maximum a posteriori 
(MAP) estimation as in \citep{tresp2001}, we instead  focus on the faster maximization-maximization (MM) strategy. For generative mixtures of GP experts, an MM algorithm was developed in \citep{chen2014precise}.

Our MM algorithm provides MAP estimation for the augmented model by optimising the augmented posterior:
\begin{align*}
	&\pi(\psi, \theta, \sigma^2, \bz, \tilde{\bof }\mid  \by, \bx, \tilde{\bx}) \propto p(\by,\bz, \tilde{\bof } \mid \bx, \tilde{\bx},\psi, \theta, \sigma^2 ) \pi(\psi, \theta, \sigma^2)\\
	&\quad = \prod_{l=1}^L \left[\prod_{i:z_i=l} w_l(x_i;\psi) \text{N}(y_i\mid  \widehat{\mu}_{l,i}, 
		\sigma^{2 }_l +\lambda_{l,i} ) \right]\text{N}(\tilde{\bof}_l \mid \bm{\mu}_l, \mathbf{K}_{l,M_l}) \pi(\psi, \theta, \sigma^2), 
\end{align*}
where $\theta =(\theta_1,\ldots, \theta_L)$, $\sigma^2 =(\sigma^2_1,\ldots, \sigma^2_L)$, and $\tilde{\bof} =(\tilde{\bof}_1,\ldots, \tilde{\bof}_L)$ collect the expert-specific GP hyperparameters, noise variance, and pseudo-targets, respectively. In the following, we consider vague priors $\pi(\psi, \theta, \sigma^2) \propto 1$, and thus equivalently optimise the augmented log posterior:
\begin{equation*}
\begin{aligned}
&\log(\pi(\psi, \theta, \sigma^2, \bz, \tilde{\bof }\mid  \by, \bx, \tilde{\bx}) ) =\text{const.}  + \sum_{l=1}^L \log(  \text{N}(\tilde{\bof}_l \mid \bm{\mu}_l, \mathbf{K}_{l,M_l}))\\ 
&\quad + \sum_{l=1}^L \sum_{i:z_i=l} \log(w_{l}(x_i; \psi)) +  \log( \text{N}(y_i \mid  \widehat{\mu}_{l,i},\sigma^{2 }_l +\lambda_{l,i})).\\ 
 \end{aligned}
  \end{equation*}
 For GP experts, this provides maximum marginal likelihood estimation or type II MLE of the GP hyperparameters and noise variance $(\theta_l, \sigma^{2}_l)$, as the GP regression functions are marginalized.  For sparse GP experts, we have the additional parameters $(\tilde{\bx}_l, \tilde{\bof}_l)$. The pseudo-inputs $\tilde{\bx}_l$ are treated as hyperparameters to be optimized, and while pseudo-targets $\tilde{\bof}_l$ can be analytically integrated, we also estimate $\tilde{\bof}_l$ with its posterior mean for faster inference (see Section \ref{sec:mm} for further details). 
The MM algorithm is an iterative conditional modes algorithm \citep{KITTLER1984} (can also be viewed as a coordinate ascent method) that alternates between optimizing the allocation variables $\bz$ and the model parameters. It is guaranteed to never decrease the log posterior of the augmented model and therefore will converge to a fixed point \citep{kurihara2009bayesian}. However, it is susceptible to local maxima, which can be alleviated with multiple restarts of random initializations.

\subsection{Maximization-maximization}
\label{sec:mm}

Our MM algorithm is defined by iterating over the following two steps: \textbf{optimize} the
(i) \textbf{latent cluster} \textbf{allocation variables} and (ii) 
\textbf{gating and expert} \textbf{parameters}
\begin{equation}\label{eq:cluster_MM}
\begin{aligned}  
&\text{(i)}~~ {\bf z} =\argmax_{{\bf z} \in \lbrace 1, \ldots, L \rbrace^N } \log \pi({\bf z} \mid \psi, \theta, \sigma^2, \tilde{\bof}, \tilde{\bx} , \bx, \by) \, ,\\
&\text{(ii)} ~~ (\psi, \theta, \sigma^2, \tilde{\bof }, \tilde{\bx}) = \argmax_{(\psi, \theta, \sigma^2, \tilde{\bof }, \tilde{\bx})} \log \pi(\psi, \theta, \sigma^2, \tilde{\bof }  \mid \tilde{\bx},  {\bf z}, \bx, \by).
 \end{aligned}
 \end{equation}

We note that while the pseudo-targets $\tilde{\bof}$ can be marginalized, the full conditional for $\bz$ in the first step of \cref{eq:cluster_MM} would be:
\begin{align*}
	 \pi({\bf z} \mid \psi, \theta, \sigma^2, \tilde{\bx} , \bx, \by) &= \int  \pi({\bf z}, \tilde{\bof} \mid \psi, \theta, \sigma^2, \tilde{\bx} , \bx, \by) d\tilde{\bof}\\
	 &\propto p({\bf z} \mid \psi, \bx) p(\by \mid \bx, \bz, \tilde{\bx}, \theta, \sigma^2),
\end{align*}
with optimal allocation:
 \begin{align}  \label{eq:optz}
{\bf z} &= \argmax_{{\bf z} \in \lbrace 1, \ldots, L \rbrace^N } \log(\pi({\bf z} \mid \psi, \theta, \sigma^2, \tilde{\bx} , \bx, \by)) \nonumber\\
&=\argmax_{{\bf z} \in \lbrace 1, \ldots, L \rbrace^N } \sum_{i=1}^N  \log(w_{z_i}(x_i; \psi)) +  \sum_{l=1}^L  \log\left( \text{N}(\by_l \mid \bm{\mu}_l, \bm{\Sigma}_l )\right)\, .
 \end{align}
Recall from \cref{eq:sparse_marg} that $\bm{\Sigma}_l$ is a full covariance matrix; thus, the second term in \cref{eq:optz} does not factorize across $i=1,\ldots, N$ and direct optimization over the space $\lbrace 1, \ldots, L \rbrace^N$  is infeasible. An iterative conditional modes algorithm could be employed which cycles over each data point $i=1,\ldots, N$,  optimizing $z_i$ based on the conditional Gaussian likelihood given the data points currently allocated to each cluster. For each data point, the  conditional Gaussian likelihood $p(y_i \mid z_i = l, \bz_{-i}, \by_{-i})$ must be computed (where $\bz_{-i}$ and $\by_{-i}$ denote $\bz$ and $\by$ with the $i^\text{th}$ element removed), requiring rank one updates to the inverse covariance matrices for every $l=1,\ldots, L$.
 
However, we can significantly reduce the computational cost by also estimating $\tilde{\bof}$. In this case, the full conditional for $\bz$ in (i) of \cref{eq:cluster_MM} factorizes across $i=1,\ldots, N$:
\begin{align*}
	 \pi({\bf z} \mid \psi, \theta, \sigma^2, \tilde{\bof}, \tilde{\bx} , \bx, \by) &\propto p({\bf z} \mid \psi, \bx) p(\by \mid \bx, \bz, \tilde{\bx}, \tilde{\bof}, \theta, \sigma^2)\\
	&= \prod_{i=1}^N w_{z_i}(x_i; \psi) \text{N}(y_i \mid  \widehat{\mu}_{z_i,i},\sigma^{2 }_{z_i} +\lambda_{z_i,i}).
\end{align*}
 Thus, the allocation can be done in parallel across the $N$ data points: 
\begin{equation}\label{eq:cluster_MM2}
	\begin{aligned}
		z_i = \argmax_{z_i \in \lbrace 1, \ldots, L \rbrace} \log(w_{z_i}(x_i; \psi)) + \log(\text{N}(y_i\mid  \widehat{\mu}_{z_i,i}, \sigma^{2 }_{z_i} +\lambda_{z_i,i})).
	\end{aligned}
\end{equation}

In the second step of the MM algorithm in \cref{eq:cluster_MM}, we perform two sub-steps, first optimizing the gating network and expert parameters with $\tilde{\bof }$ marginalized, and then, given those optimal values, estimating $\tilde{\bof }$:
\begin{equation}\label{eq:cluster_MM3}
	\begin{aligned}  
			&\text{(a)} ~~ (\psi, \theta, \sigma^2,  \tilde{\bx}) = \argmax_{(\psi, \theta, \sigma^2, \tilde{\bx})} \log \pi(\psi, \theta, \sigma^2  \mid \tilde{\bx},  {\bf z}, \bx, \by),\\
		&\text{(b)}~~  \tilde{\bof } = \argmax_{\tilde{\bof }} \log \pi(\tilde{\bof }  \mid \tilde{\bx},  {\bf z}, \bx, \by, \psi, \theta, \sigma^2) .
	\end{aligned}
\end{equation}
 In (a) of \cref{eq:cluster_MM3}, the full conditional of $(\psi, \theta, \sigma^2)$ is:
 	\begin{align*}
 		\pi(\psi, \theta, \sigma^2  \mid \tilde{\bx},  {\bf z}, \bx, \by) &\propto p(\bz \mid \bx, \psi) p(\by \mid  \bx, \tilde{\bx}, \bz, \tilde{\bof },  \theta, \sigma^2 ) \\
 		&= \prod_{l=1}^L \left[\prod_{i:z_i=l} w_l(x_i;\psi) \right] \text{N}(\by_l\mid  \bm{\mu}_l, \bm{\Sigma}_l).
 	\end{align*}
  Thus, optimization of the gating network and expert parameters can be done in parallel, both between 
 each other as well as across $l=1,\ldots, L$ for the experts. 
 Specifically, the optimal gating network are:
 \begin{equation}\label{eq:classification_MM}
 	\begin{aligned}
 		\psi 
 		= \argmax_{\psi \in \Psi}\sum_{l=1}^L\sum_{i: z_i=l}  h_{l}(x_i; \psi) - \sum_{i=1}^N \log\left( \sum_{l=1}^L \exp(h_l(x_i ; \psi)) \right),
 	\end{aligned}
 \end{equation}
and expert parameters (GP hyperparameters, noise variance, and pseudo-inputs) are estimated by optimizing the log marginal likelihood in \cref{eq:sparse_marg}:
 \begin{equation} \label{eq:regress_MM}
 	\begin{aligned}
 		(\theta_l, \sigma^{2}_l , \tilde{\bx}_l) =  \argmax_{\theta \in \Theta, \sigma^2 \in \mathbb{R}_+, \tilde{\bx} \in \mathbb{R}^{M_l \times d} }  
 		\log\left( \text{N}(\by_l \mid \bm{\mu}_l, \bm{\Sigma}_l )\right)\, .
 	\end{aligned}
 \end{equation}
Then, in (b) of \cref{eq:cluster_MM3}, the full conditional for $\tilde{\bof}_l$ is: 
\begin{align*}	
	\pi(\tilde{\bof }_l  \mid \tilde{\bx}_l, \bx, \by, \theta, \sigma^2) &\propto  	p(\tilde{\bof }_l  \mid \tilde{\bx}_l, \theta, \sigma^2) 	p(\by \mid \tilde{\bof }_l  , \tilde{\bx}_l, \bx, \by, \theta, \sigma^2)\\
	&= \text{N}(\tilde{\bof}_l \mid \bm{\mu}_l, \mathbf{K}_{l,M_l}) \prod_{i:z_i=l} \text{N}(y_i\mid  \widehat{\mu}_{l,i}, \sigma^{2 }_{l} +\lambda_{l,i}),
\end{align*}
which, following standard derivations, is shown to be Gaussian with covariance matrix $\mathbf{K}_{l,M_l} (\mathbf{Q}_{l,M_l} )^{-1}   \mathbf{K}_{l,M_l}$ and mean:
\begin{align}
&\text{E}[ \tilde{\bof}_l \mid  \theta_l, \sigma^{2}_l, \tilde{\bx}_l, \by_l, \bx_l ] \nonumber \\
&\quad=\bm{\mu}_l + \mathbf{K}_{l,M_l} (\mathbf{Q}_{l,M_l} )^{-1} \left(\mathbf{K}_{l, M_l N_l} ( \bm{\Lambda}_l +\sigma^{2}_l \mathbf{I}_{N_l})^{-1}(\by_l- \bm{\mu}_l)\right), \label{eq:ftilde_pmean}
 \end{align}
 where $\mathbf{Q}_{l,M_l} = \mathbf{K}_{l,M_l} + \mathbf{K}_{l,M_l N_l} ( \bm{\Lambda}_l +\sigma^{2}_l \mathbf{I}_{N_l})^{-1} (\mathbf{K}_{l,M_l N_l} )^T$. 
 Thus, the optimal  $\tilde{\bof}_l$ in (b) of \cref{eq:cluster_MM2} is the posterior mean in \cref{eq:ftilde_pmean}.

\subsection{A fast approximation: CCR}
\label{sec:ccr}

The MM algorithm iterates between \textbf{clustering} in  \cref{eq:cluster_MM2} and in parallel \textbf{classification} in \cref{eq:classification_MM} and \textbf{regression} in \cref{eq:regress_MM,eq:ftilde_pmean}. This closely resembles the CCR algorithm recently introduced in \citep{EtienamLaw}. 
The important differences are that 1) CCR is a one pass algorithm that does not iterate between the steps and 2)  CCR approximates the clustering in the first step of the MM algorithm by 
a) careful re-scaling the data to emphasize the output $y$ in relation to $x$ and 
b) subsequently applying a fast clustering algorithm, e.g. K-means \citep{hartigan1979}, Gaussian mixture model  \citep[GMM,][]{mclachlan1988mixture}, or DB-scan \citep{ester1996}, 
to jointly cluster the rescaled $(y,x)$. 
 We also note that the original formulation of CCR performs an additional clustering step 
 so that the allocation variables 
 used by the regression correspond to the prediction of the classifier;
 this is equivalent to the clustering step of the MM algorithm in \cref{eq:cluster_MM2} 
including only 
the term associated to the gating network.
 
 This novel connection sheds light on the MoE model underlying the CCR algorithm and
  allows us to view  CCR as a fast, one-pass approximation to the MM algorithm for MoEs. 
Therefore, we can use CCR to construct a fast approximation of the proposed deep mixture of sparse GP experts. Moreover, this connection can be used to obtain a fast approximation to other discriminative MoE architectures.   
As shown in the following section, CCR provides a good, fast approximation for many numerical  examples. 
If extra computational resources are available, 
 the CCR solution can be improved through additional MM iterations 
 (i.e. it provides a good initialization for the MM algorithm). 
However, in our examples, we notice that the potential for further improvement is limited. 
We find that MM with random initialization can also produce a reasonable estimate in many examples after two iterations; 
we refer to this algorithm as MM2r. 
It is fast, but we will see that it takes approximately 2-3 times longer than CCR.

\subsection{Complexity considerations}
\label{sec:complexity}

Suppose we parameterize the DNN with $p_c$ parameters, 
and each of the sparse GP experts is approximated with $M_l$ pseudo-inputs. 
The MM algorithm described in \cref{sec:mm} incurs a cost per iteration 
of clustering the $N$ points given the current set of parameters \cref{eq:cluster_MM}.
This cost is $\cO(N\sum_{l=1}^L M_l^2)$, and it is parallel in $N$. 
The algorithm also incurs a cost per iteration 
of classification with 
$N$ points and $L$ regressions using $N_1, \dots, N_L$ points, 
where $N= \sum_{l=1}^LN_l$ in \cref{eq:cluster_MM}.
These operations can also be done in parallel. 
The cost for the classification is $\cO(Np_c)$ assuming that the 
number of epochs for training is $\cO(1)$.
The cost for the regressions is $\cO(\sum_{l=1}^L M_l^2 N_l)$.
Hence the total cost for \cref{eq:cluster_MM} is $\cO(N p_c + \sum_{l=1}^L M_l^2 N_l)$,
which can be roughly bounded by $\cO(N P_{\rm max})$, 
where $P_{\rm max} = {\rm max}\{p_c, M_1^2,\dots, M_L^2\}$.
Randomly initialized MM cannot be expected to provide reasonable results 
after one pass, however with sparse GP models the first iteration provides a significant
improvement which is also sometimes reasonable. 
Ignoring parallel considerations, the total cost for 2-pass MM (MM2r) is 
$\cO(2 N p_c + \sum_{l=1}^L M_l^2 (2N_l+N))$.

For CCR, the cost is the same for the second step \cref{eq:cluster_MM}, 
while the first step is replaced with a chosen clustering algorithm, e.g. GMM, which incurs a cost of $\cO(NL)$. 
The latter iterates between steps which can be parallelized in different ways.
The total cost of CCR is hence $\cO(N (p_c+L) + \sum_{l=1}^L M_l^2 N_l) = \cO(N P_{\rm max})$.
This is a {\em one pass} algorithm, which often provides acceptable results.
The overhead for MM2r vs. CCR for our model is then roughly $\cO((P_{\rm max}-L)N)$. 
More precisely it is $\cO(N (p_c-L) + \sum_{l=1}^L M_l^2 (N+N_l) )$.

\subsection{Prediction}
\label{sec:prediction}

There are two approaches that can be employed to predict $y^*$ at a test value $x^*$. \textbf{Hard allocation based prediction} 
is based on the single best regression/expert: 
\begin{align} \label{eq:hard} 
z^* =\argmax_{z^* \in \lbrace 1, \ldots, L \rbrace} \log(w_{z^*}(x^*; \psi)) \, , \,\,\, y^* = \widehat{\mu}_{z^*}(x^*) \, ,
 \end{align}
where (similiar to \cref{eq:sparseGP_pred}), the GP expert's prediction is:
\begin{align*}
	\widehat{\mu}_{z^*}(x^*) = 	\mu_l + (\mathbf{k}_{l,M_l,x^*})^T(\mathbf{K}_{l,M_l})^{-1} (\tilde{\bof}_l- \bm{\mu}_l),
\end{align*}	
with $\mathbf{k}_{l,M_l,x^*}$ denoting the vector of length $M_l$ with elements $K_{\phi_l}(\tilde{x}_{l,j}, x^*)$. 
Whereas \textbf{soft allocation based prediction} is given by a weighted average
\begin{equation}\label{eq:soft}
y^* = \sum_{l=1}^L w_{l}(x^*; \psi) \widehat{\mu}_{l}(x^*) \, .
\end{equation}
Soft-allocation may be preferred in cases when there is not a clear jump in the unknown function, thus allowing us to smooth the predictions in regions where the classifier is unsure.  Similarly, the variance or density of the output or regression function can also be computed at any test location to obtain measures of uncertainty in our predictions. 

\textbf{Hard allocation based density estimation} delivers 
a Gaussian approximation for a test value $x^*$, 
 $$\widehat{p}(y\mid x^*, z^*=l) =  \text{N}\left(y\mid \widehat{\mu}_{l}(x^*),\widehat{\sigma}^{2}_l(x^*) \right) \, ,$$
with allocation $z^*$ as in \cref{eq:hard}, and where $ \widehat{\sigma}^{2}_l(x^*) = \lambda_{l, x^*} + \sigma^{2}_l + (\mathbf{k}_{l,M_l,x^*})^T (\mathbf{Q}_{l,M_l})^{-1} \mathbf{k}_{l,M_l,x^*}$ and $ \lambda_{l,x^*}$ is computed as in \cref{eq:sparseGP_pred2} at $x^*$. 
\textbf{Soft allocation based density estimation} delivers a mixture of Gaussians for a test value $x^*$, 
similar to \cref{eq:soft}:
$$\widehat{p}(y\mid x^*) = \sum_{l=1}^L w_{l}(x^*; \psi) \text{N}(y\mid \widehat{\mu}_l(x^*),\widehat{\sigma}^{2}_l(x^*) ).$$
In cases when the density of the output may be multi-modal, looking at point predictions alone is not useful. In this setting, the soft density estimates are preferred, allowing one to capture and visualise the multi-modality. 

\section{Numerical Experiments}
\label{sec:experiments}
In this section, we perform a range of experiments to highlight the flexibility of our model and compare the accuracy and speed of the 
MM and CCR algorithms. 
For the sparse GP experts, we use the isotropic squared exponential covariance function with a variable number of inducing points $M_l$ based on the cluster sizes \citep{burt2020,nieman2021contraction}. The pseudo-inputs $\tilde{\bx}^l$ are initialized via K-means and the GP hyperparameters are initialized based on the scale of the data. 
The number of experts $L$ is determined apriori using the Bayesian information criterion (BIC) to compare the GMM clustering solutions of the rescaled $(y,x)$ across different values of $L$. The choice of GMM is motivated by allowing elliptically-shaped clusters in the rescaled space, while keeping cost low (e.g. if compared to the K-means).  In the Appendix \ref{app:numexperts}, we instead consider using cross-validation and examine the log-likelihood on the held-out data to select the number of components, but we found that this significantly increases run time compared to BIC, with similar performance metrics. A table reporting the number of experts can also be found in Appendix \ref{app:numexperts}.
The architecture of the DNN gating network is chosen with the use of Auto-Keras~\citep{jin2019auto} 
and a quadratic regularization is used 
with a value of $0.0001$ for the penalization parameter. 
The adaptive stochastic gradient descent solver Adam \citep{kingma2014adam}
is used to optimize the model weights and biases, with 
a validation fraction of $0.1$ 
and 
a maximum of $500$ epochs.
The GPy package \citep{gpy2014} and Keras \citep{chollet2015keras} were used to train the experts and gating network respectively.

For each experiment we performed 5-fold cross-validation with random shuffling of the data. The experiments are executed on Intel Core i7 (I7-7820HQ) with 4 cores and with 16GB of RAM.

\subsection{Simulations}\label{sec:simulations}
As a motivating toy example, we generate inputs within each cluster $l = 1, 2$ from a noisy 2D spiral and outputs $y = (2l -1) (x_1^2 + x_2^2) + \epsilon $. \cref{fig:toy_example} demonstrates that flexible DNN gating networks, over quadratic classifiers (e.g. logistic regression (LR)), are required to recover the true allocations. When combined with GP experts, this leads to improved accuracy ($R^2=98.74\%, \, 86.98\%, \, 97.04\%$ for DNN+GP, LR+GP, MDN, respectively) and tighter credible intervals (CI) that maintain the desired coverage (average CI length = $0.25,\, 1.36,\, 0.41$ and empirical coverage (EC) = $95\%,\, 100\%,\, 98.7\%$ for DNN+GP, LR+GP, MDN, respectively).
In particular, for outlying test points (red squares in \cref{fig:toy_example}), MDN provides highly overconfident inaccurate predictions, whereas predictions and CIs are more accurate and realistic for the proposed DNN+GP.

\subsection{Datasets}\label{sec:datasets}

Our experiments range from small to large datasets of varying dimensions and complexity and mainly aim to highlight the model's capability to flexibly estimate the conditional density and 
capture issues such as discontinuity, non-stationarity, heteroscedasticity, and multi-modality that challenge standard GP models. First, the Motorcycle dataset \citep{silverman1985} consists of $N=133$ measurements with $d=1$. Second, the NASA  dataset \citep{gramacy2008bayesian} comes from a computer simulator of a NASA rocket booster vehicle with $N=3167$; we focus on modelling the lift force as a function of the speed (mach), the angle of attack (alpha), and the slide-slip angle (beta), i.e. $d=3$.
Our third experiment is the Higdon function \citep{higdon2002, gramacy2008bayesian};  $\cX=[0,20]$ and
$N=1000$. 
Next, $N=10,000$ points are generated from the Bernholdt function \citep{EtienamLaw}; in this case, $\cX=[-4,10]^2$ and $f(x) = g(x_1)g(x_2)$, 
where $g(x)$ is the piece-wise smooth 
function studied in  \citep{monterrubio2018posterior}. 
The kin40k dataset \citep{seeger03a} is a popular benchmark for GP regression methods and consists of $N = 40,000$ points that describe the location of a robotic arm as a highly nonlinear function of control input with $d = 8$.

\subsubsection{$\chi$ dataset}\label{sec:chi_data}

A tokamak is a device which uses magnetic fields to
confine hot plasma in the shape of a torus. It is the leading
candidate for production of controlled thermonuclear power,
for use in a prospective future fusion reactor \citep{tokamakweb}.

One dimensional radial transport modeling \citep{park2017} using theory-based models such as GLF23 \citep{waltz1997gyro}, MMM95 \citep{bateman1998predicting}, and TGLF \citep{staebler2007theory} plays an essential role in interpreting experimental data and guiding new experiments for magnetically confined plasmas in tokamaks. Turbulent transport resulting from micro-instabilities have a strong nonlinear dependency on the temperature and density gradients. One of the key characteristics is a sharp increase of turbulent flux as the gradient of temperature increases beyond a certain critical value.
This leads to a highly nonlinear and discontinuous function of the inputs.

\begin{figure}[tbh]
    \centering
    \includegraphics[width=0.81\textwidth]{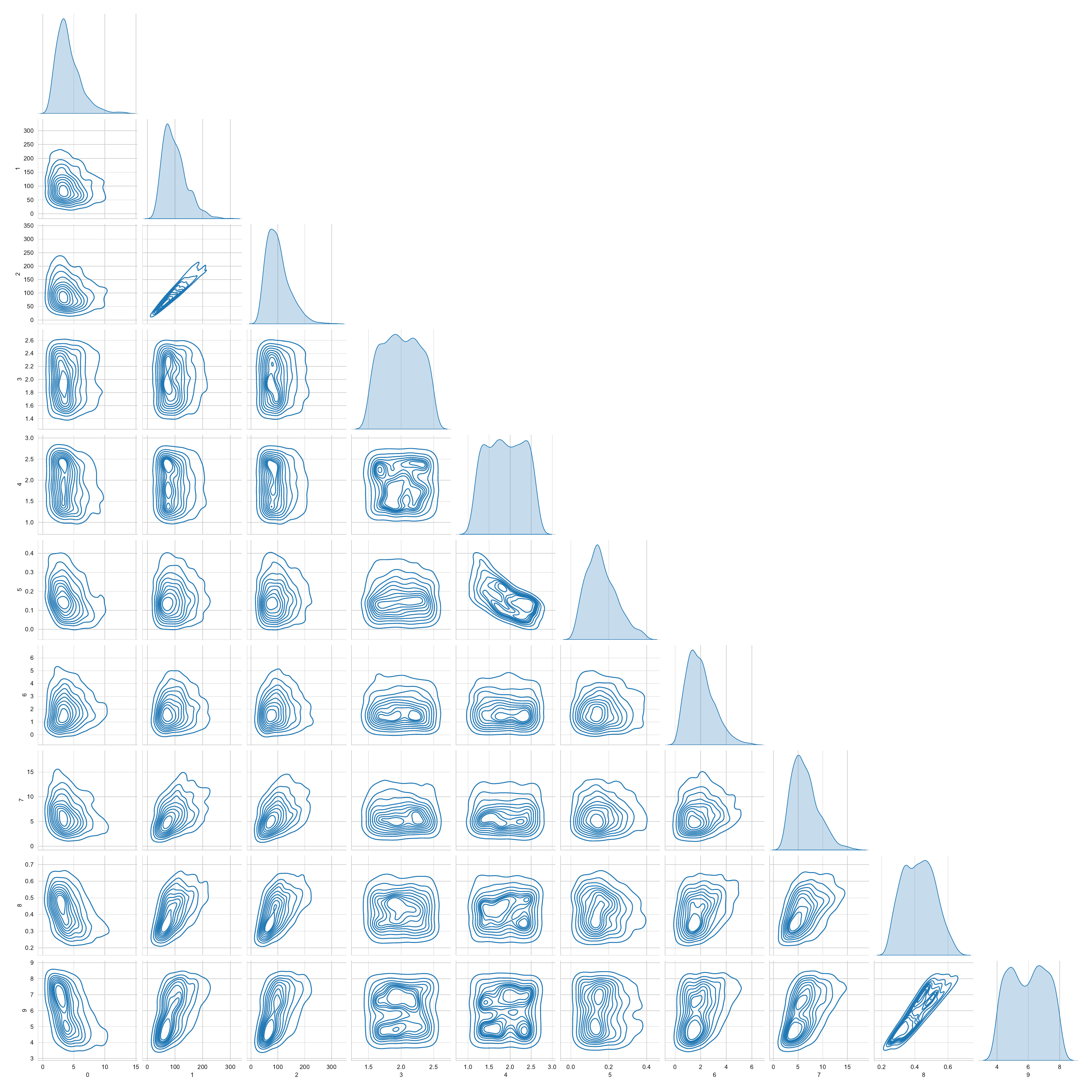}
    \caption{Input distribution marginals of a $\chi$ dataset}
    \label{fig:chi_input}
\end{figure}

Here we consider an analytical stiff transport model that describes turbulent ion energy transport in tokamak plasmas \citep{EtienamLaw, janeschitz20021}:
\begin{equation}\label{eq:chi}
\chi = S (R T'/T - (R T'/T)_{\rm crit})^\alpha
H\left (\left\vert \frac{(R T'/T)}{(R T'/T)_{\rm crit}} \right\vert - 1\right) \, ,
\end{equation}
where $\chi$ 
is ion thermal diffusivity, $H(\cdot)$ is the Heaviside function,
$R$ is the major radius,
and $T'$ is the radial derivative of ion temperature.
The normalized critical gradient
$(R T'/T)_{\rm crit}$ of ion temperature is calculated using IFS/PPPL model \citep{kotschenreuther1995quantitative},
which is a nonlinear function of electron density ($n_e$), electron and ion temperatures ($T_e,T$), safety factor ($q$), magnetic shear ($\hat{s}$), effective charge ($Z_{\rm eff}$)
and the normalized gradient of ion temperature $(RT'/T)$ and density $(Rn'/n)$. It is assumed that $S=1$ and $\alpha=1$.
Considering $(T, T')$ and $(n, n')$ as two input parameters
each, this gives a total of 10 inputs $x \in \bbR^{10}$.
The output \eqref{eq:chi} is $y \in \bbR_+$.
Basic primitive model inputs are chosen uniformly at
random from a hypercube $\omega \in [0,1]^{17}$,
which then give rise to realistic inputs $x \in \bbR^{10}$,
which concentrate on a manifold in the ambient space.
See \citep{kotschenreuther1995quantitative} for details
of the model, and Figure \ref{fig:chi_input} for visualization
of the input distribution histogram.

The data consists of $N=150,000$ simulations from \eqref{eq:chi}.
This model presents a challenge for the competing MoE methods, 
due to input dimensionality and data size, 
and for competing PoE methods, due to the sharp nonlinearities 
and discontinuities.
It therefore serves to highlight the value of our method.
Results appear in the bottom row of the tables.

\subsection{Results}

For our model, we compare the MM algorithm 
with the fast CCR and MM2r approximations. The MM algorithm is initialized at the CCR solution and iterates until the reduction in the $R^2$ is below a threshold of $0.0001$ or a maximum number of iterations is reached. 
Competing models are two product of GP experts models, 
gPoE \citep{cao2014}, RBCM \citep{deisenroth2015},
a mixture of GP experts, FastGP \citep{nguyen2014}, 
and MDN \citep{bishop1994}, as well as a Bayesian treed-based model \citep[BART][]{bart}. Treed GP models \citep{gramacy2008bayesian} were also considered, although the cost was so much larger that the method is not competitive (see \cref{tab:tgp}). We also report results for other methods, including ORTHNAT \citep{orthnat}, PPGPR \citep{ppgpr}, DSPP \citep{dspp}, and Deep GPs \citep{deepgp}.

First, we observe that CCR has similar or improved test accuracy compared to MM2r (\cref{tab:oneb}), and CCR reduces wall-clock time (\cref{tab:two}) by a factor of 2-3 for all experiments. The CCR solution can be further refined through MM iterations (\cref{tab:oneb}); however this comes at a higher computational cost. Moreover, in practice we observe little to no improvement in accuracy with further MM iterations, across all datasets.  

Compared with state-of-the-art GP, neural network, and tree-based  benchmarks,  
our model has the highest test accuracy in almost all the experiments considered;
the one exception is RBCM for Motorcycle, where the accuracy is slightly higher, but well within the standard deviation.
For the smaller datasets, 
CCR is slightly slower than the PoE models (gPoE and RBCM) and FastGP, 
however this is offset by the improved accuracy.
CCR is substantially faster than all competitors for the $\chi$ model, with $d=10$ and (relatively) big data $N=150,000$.

\begin{table*}[t]
\centering
\caption{$R^2$ test accuracy (\%) on the five datasets for our model with CCR, CCR-MM, and MM2r algorithms and MDN, gPoE, RBCM, and FastGP benchmarks.  Results obtained with 5-fold cross-validation (standard deviation in brackets).}
\resizebox{\textwidth}{!}{ 
\begin{tabular}{ | l | c | c | c || c | c | c | c | c | c | c | c | c |}
\hline
\textbf{Model}          	& \textbf{CCR}			& \textbf{CCR-MM} 		& \textbf{MM2r} 	& \textbf{MDN}		& \textbf{gPoE} 	& \textbf{RBCM} 		& \textbf{FastGP}	& \textbf{BART} 	& \textbf{ORTHNAT}	& \textbf{PPGPR}	& \textbf{DSPP}	& \textbf{Deep GP} \\ 
\hline
\textbf{Motorcycle}	& 76.70 (9.24)  			& 77.84 (9.87)    		& 74.60 (12.03)   	& 71.27 (10.84) 	& 74.46 (8.12) 		& \textbf{78.46 (8.42)} 	& 74.94 (10.32) 	& 75.12 (8.65)		& 78.18 (10.39) 		& 77.27 (9.27) 	& 75.82 (11.51)	& 74.09 (10.04) \\
\textbf{NASA }          	& \textbf{97.07 (1.39)}  	& 96.94 (1.50)    		& 95.75 (1.49)    	& 96.80 (1.82)  		& 93.97 (2.01) 		& 94.52 (1.13) 			& 94.71 (1.42)    	& 96.72 (1.93) 	& 95.96 (1.70) 		& 96.85 (1.22) 	& 94.82 (1.95)		& 96.74 (1.62) \\
\textbf{Higdon }        	& 99.91 (0.03)  	    		& \textbf{99.96 (0.01)}    	& 99.89 (0.02)    	& 99.35 (0.02)  		& 99.90 (0.02) 		& 99.94 (0.02) 			& 99.87 (0.02) 		& 99.52 (0.03) 	& 99.44 (0.03) 		& 99.71 (0.04) 	& 99.45 (0.05)		& 99.94 (0.02) \\
\textbf{Bernholdt }     	& \textbf{99.50 (0.40)}  	& 98.00 (0.58)     		& 94.81 (0.93)     	& 93.34 (3.05)  		& 89.69 (2.23) 		& 94.71 (2.15) 			& 95.90 (0.29) 		& 94.17 (0.62)  	& 91.25 (1.06) 		& 90.98 (1.78) 	& 96.39 (1.08)		& 97.12 (0.68) \\

\textbf{kin40k} 		& 94.53 (1.21)  			& \textbf{95.02 (1.32)}    	&  92.38 (2.03)   	& 90.75 (2.85) 		& 92.08 (1.97) 		& 91.36 (2.41) 			& 92.94 (1.78)  		& 94.72 (1.99)  	& 93.91 (2.13) 		& 93.06 (2.41) 	& 91.62 (2.12)		& 91.68 (2.06) \\

$\chi_{\text{150k}}$ 	& 95.71 (0.92)  			& \textbf{97.53 (1.29)}    	& 93.62 (2.74)    	& 89.90 (4.84)  		& 91.92 (1.18) 		& 90.71 (9.87) 			& 92.99 (2.42)  		& 91.47 (2.81)  			& 92.38 (4.38) 				& 94.44 (3.85)			& 92.04 (1.57)			& 92.75 (2.91) \\
\hline
\end{tabular}}
\label{tab:oneb}\end{table*}

\begin{table*}[t]
\centering
\caption{Mean wall-clock time (in seconds) on the five datasets for our model with CCR, CCR-MM, and MM2r algorithms and MDN, gPoE, RBCM, and FastGP benchmarks. Results obtained with 5-fold cross-validation.}
\resizebox{\textwidth}{!}{ 
\begin{tabular}{ | l | c | c | c || c | c | c | c | c | c | c | c | c | c |}
\hline
\textbf{Model}      	& \textbf{CCR} 	& \textbf{CCR-MM}	& \textbf{MM2r}		& \textbf{MDN}		& \textbf{gPoE}		& \textbf{RBCM}	& \textbf{FastGP}	& \textbf{BART}	& \textbf{ORTHNAT}	& \textbf{PPGPR}	& \textbf{DSPP}	& \textbf{DeepGP} \\ 
\hline
\textbf{Motorcycle} 	& 4.7         	& 36.3        		& 11.2      			& 21.6    			& \textbf{4.2} 		& 5.5           		& 7.8				& 8.7 			& 6.9 				& 4.3			& 5.3			& 10.9 \\
\textbf{NASA }      	& 10.1       	& 25.8        		& 20.4        		& 188.2  			& 9.4 			& \textbf{8.5}            	& 10.3			& 27.3 			& 10.0 				& 27.4			& 23.2			& 10.4 \\
\textbf{Higdon }    	& 9.7        		& 46.8        		& 22.9         		& 82.1    			& 7.2 			& \textbf{6.9}             	& 7.7				& 14.7 			& 13.6 				& 11.5			& 8.1			& 9.5 \\
\textbf{Bernholdt } 	& 66.3        	& 232.6      		& 159.5       		& 252.7  			& \textbf{65.5} 		& 77             		& 69.1			& 77.1 			& 67.2 				& 80.4			& 84.2			& 79.6 \\
\textbf{kin40k} 		& \textbf{85.7}  	& 301.4    			&  179.2   			& 284.9  			& 92.3 			& 127.1 			& 120.8			& 107.1 			& 119.4 				& 116.9			& 101.3			& 91.6 \\
$\chi_{\text{150k}}$ 	& \textbf{495.7}	& 1290.1     		& 1003.5     		& 1886.4    		& 1711.9 			& 1542.5 			& 1170.9			& 1650.8 		& 1076.9 			& 1475.3			& 1620.4			& 1511.5 \\
\hline
\end{tabular}}
\label{tab:two}\end{table*}

\begin{table*}[t]
\centering
\caption{Empirical coverage ($\text{EC}_{95}$) and average length of 95\% CIs ($\bar{\text{CI}}_{95}$), averaged over 5-folds. We highlighted the model that has the shortest average length of 95\% CIs while providing empirical coverage $\geq 95\%$.}
\resizebox{\textwidth}{!}{ 
\begin{tabular}{ | l | c  c || c  c | c  c | c  c | c  c | c  c | c c | c c | c c | c c |}
\hline
\multirow{2}{*}{\textbf{Model}}	& \multicolumn{2}{c||}{\textbf{CCR}} 			& \multicolumn{2}{c|}{\textbf{MDN}}			& \multicolumn{2}{c|}{\textbf{gPoE}} 			& \multicolumn{2}{c|}{\textbf{RBCM}} 		& \multicolumn{2}{c|}{\textbf{FastGP}}		& \multicolumn{2}{c|}{\textbf{BART}}			& \multicolumn{2}{c|}{\textbf{ORTHNAT}}						& \multicolumn{2}{c|}{\textbf{PPGPR}}  	& \multicolumn{2}{c|}{\textbf{DSPP}} 				& \multicolumn{2}{c|}{\textbf{DeepGP}} \\ 
\cline{2-21}
						& $\text{EC}_{95}$ & $\bar{\text{CI}}_{95}$ 	& $\text{EC}_{95}$ & $\bar{\text{CI}}_{95}$ 	& $\text{EC}_{95}$ & $\bar{\text{CI}}_{95}$ 	& $\text{EC}_{95}$ & $\bar{\text{CI}}_{95}$ 	& $\text{EC}_{95}$ & $\bar{\text{CI}}_{95}$	& $\text{EC}_{95}$ & $\bar{\text{CI}}_{95}$	& $\text{EC}_{95}$ & $\bar{\text{CI}}_{95}$			& $\text{EC}_{95}$ & $\bar{\text{CI}}_{95}$ 		& $\text{EC}_{95}$ & $\bar{\text{CI}}_{95}$					& $\text{EC}_{95}$ & $\bar{\text{CI}}_{95}$\\  \hline	
\textbf{Motorcycle}    			& \textbf{96.35} & \textbf{0.84}  				& 84.75 & 0.67 							& 98.78 & 1.18 	        						& 96.29 & 0.97 	    						& 99.46 & 1.31							& 95.17 & 1.02							& 98.22 & 1.15									& 94.23 & 1.01				& 98.70 & 0.72				& 96.86 & 1.19 \\
\textbf{NASA}    	   		& \textbf{98.38} & \textbf{0.35} 	        			& 96.94 & 0.39    						& 97.92 & 0.58 							& 88.38 & 0.43 							& 97.89 & 0.49							& 97.31 & 0.54							& 96.08 & 0.50									& 97.00 & 0.54 				& 97.14 & 0.43				& 95.85 & 0.51 \\
\textbf{Higdon}        			& \textbf{97.32} & \textbf{0.06}   			& 95.90 & 0.09 							& 98.60 & 0.10 	        						& 71.70 & 0.08  						& 99.90 & 0.09 							& 96.79 & 0.07 							& 97.36 & 0.12									& 94.56 & 0.08 				& 96.35 & 0.07 				& 94.75 & 0.11 \\
\textbf{Bernholdt}     			& \textbf{98.34} & \textbf{0.29}           		& 95.35 & 0.39      						& 96.73 & 0.43 							& 92.19 & 0.41							& 98.14 & 0.49 							& 94.24 & 0.35 							& 97.07 & 0.57									& 98.15 & 0.46 				& 94.76 & 0.51 				& 96.58 & 0.48 \\
\textbf{kin40k} 			     	& \textbf{95.12}  & \textbf{0.51}   			& 94.89 & 0.56  						& 93.67 & 0.61 							& 93.23 & 0.58 							& 94.40 & 0.66 							& 96.25 & 0.63 							& 96.24 & 0.73									& 94.92 & 0.67 				& 95.09 & 0.62 				& 94.30 & 0.67 \\
$\chi_{\text{150k}}$   		& \textbf{97.51} & \textbf{0.63}           		& 96.89 & 0.72      						& 96.44 & 0.74 							& 79.27 & 0.68							& 94.70 & 0.71 							& 95.69 & 0.87 							& 97.85 & 0.88									& 93.65 & 0.66 				& 94.33 & 0.86 				& 92.36 & 0.63 \\
\hline
\end{tabular}}\label{tab:three}
\end{table*}

\begin{figure}[htb]
	\centering
	\subfloat[Heat map of the conditional density]{\includegraphics[width=0.45\textwidth]{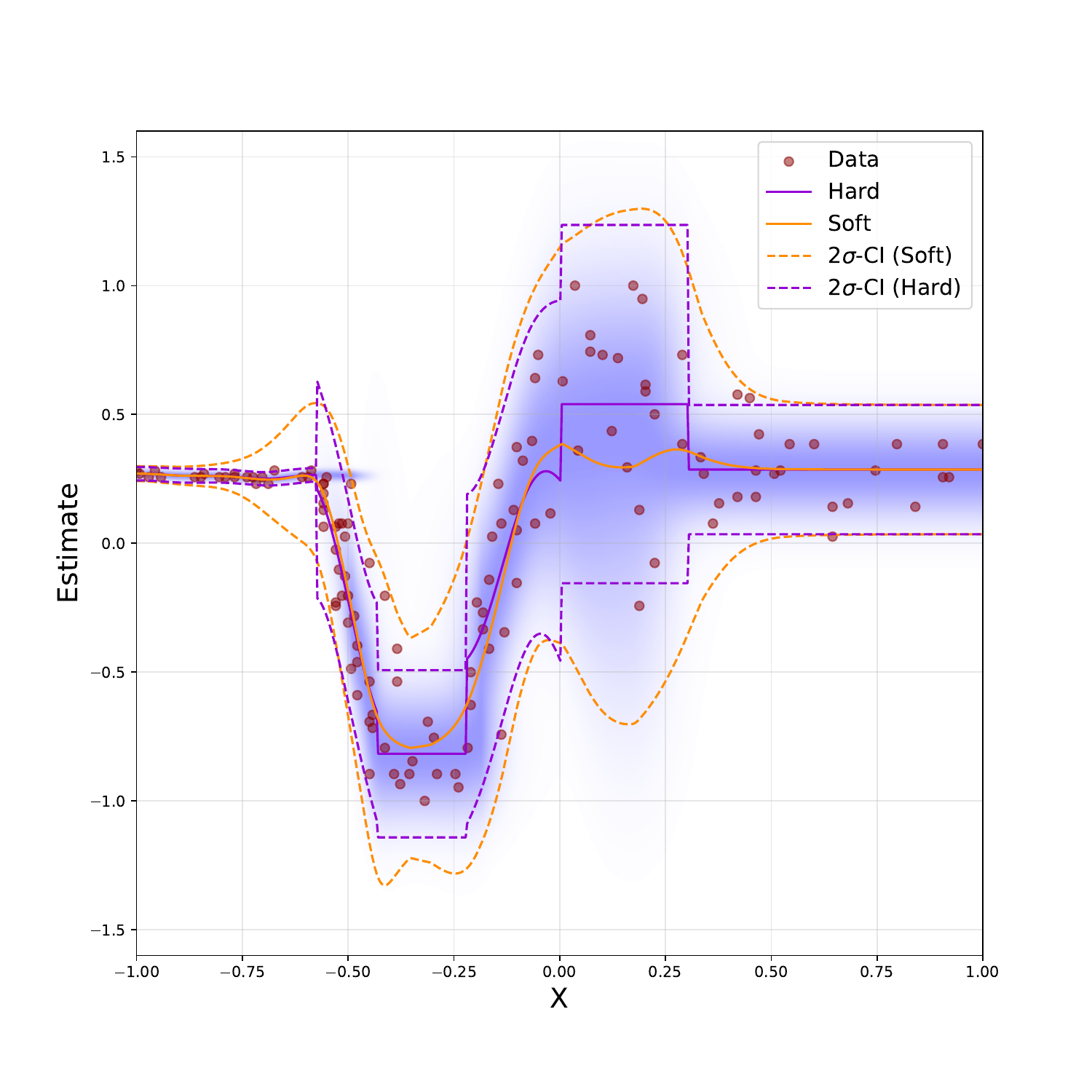}}
	\subfloat[Density estimate given $x^* = -0.478$.]{\includegraphics[width=0.45\textwidth]{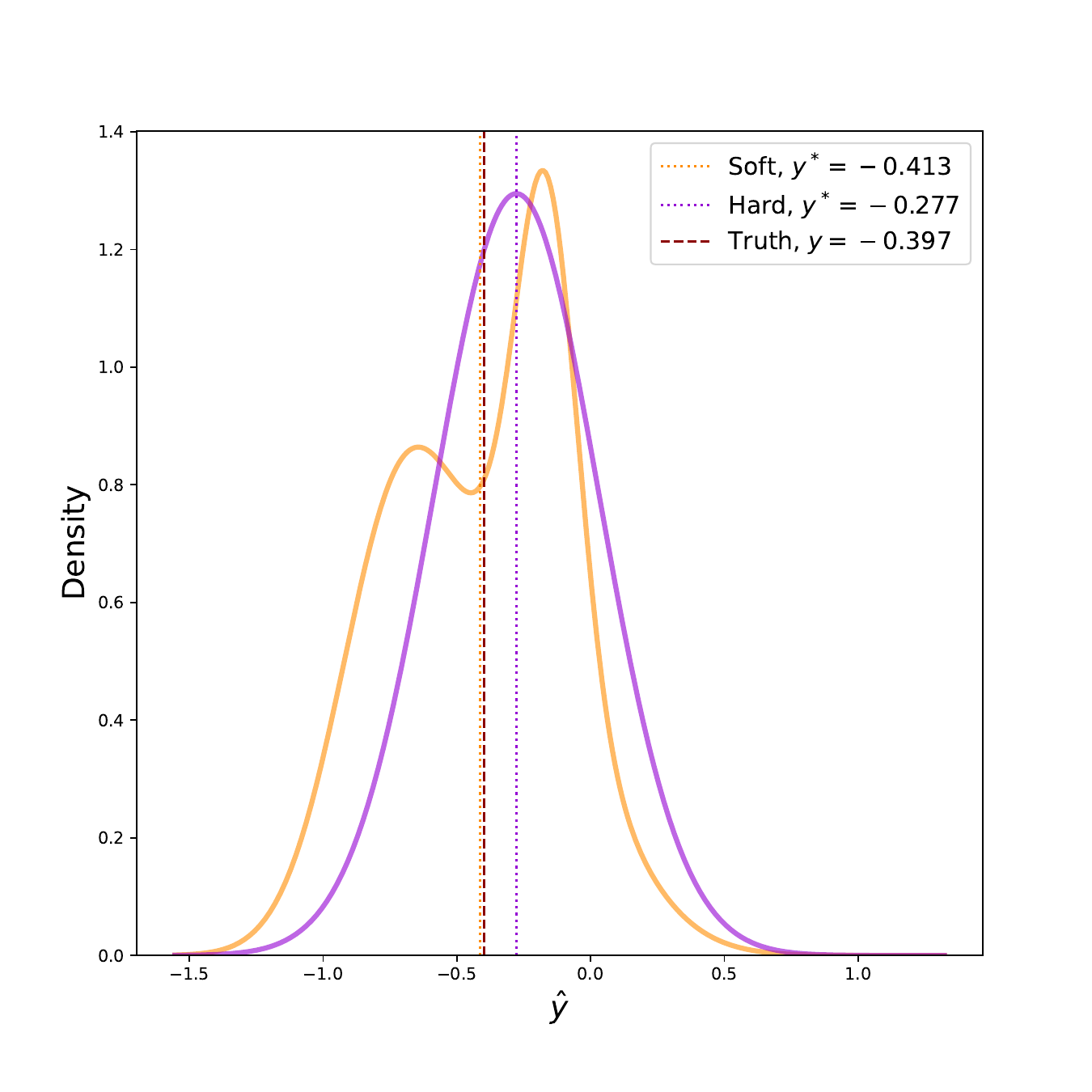}}
	\caption{ (a) Predictions (based on soft and hard allocations) with our model for the Motorcycle dataset, with two standard deviations and soft allocation based density estimates; (b) a slice representing the density estimate given  $x^* = -0.478$.}\label{fig:moto_uq}
\end{figure}

\begin{figure}[!h]
	\centering
	\subfloat[CCR]{\includegraphics[width=0.27\textwidth]{figs/motorcycle_UQ.pdf}}
	\subfloat[MDN]{\includegraphics[width=0.27\textwidth]{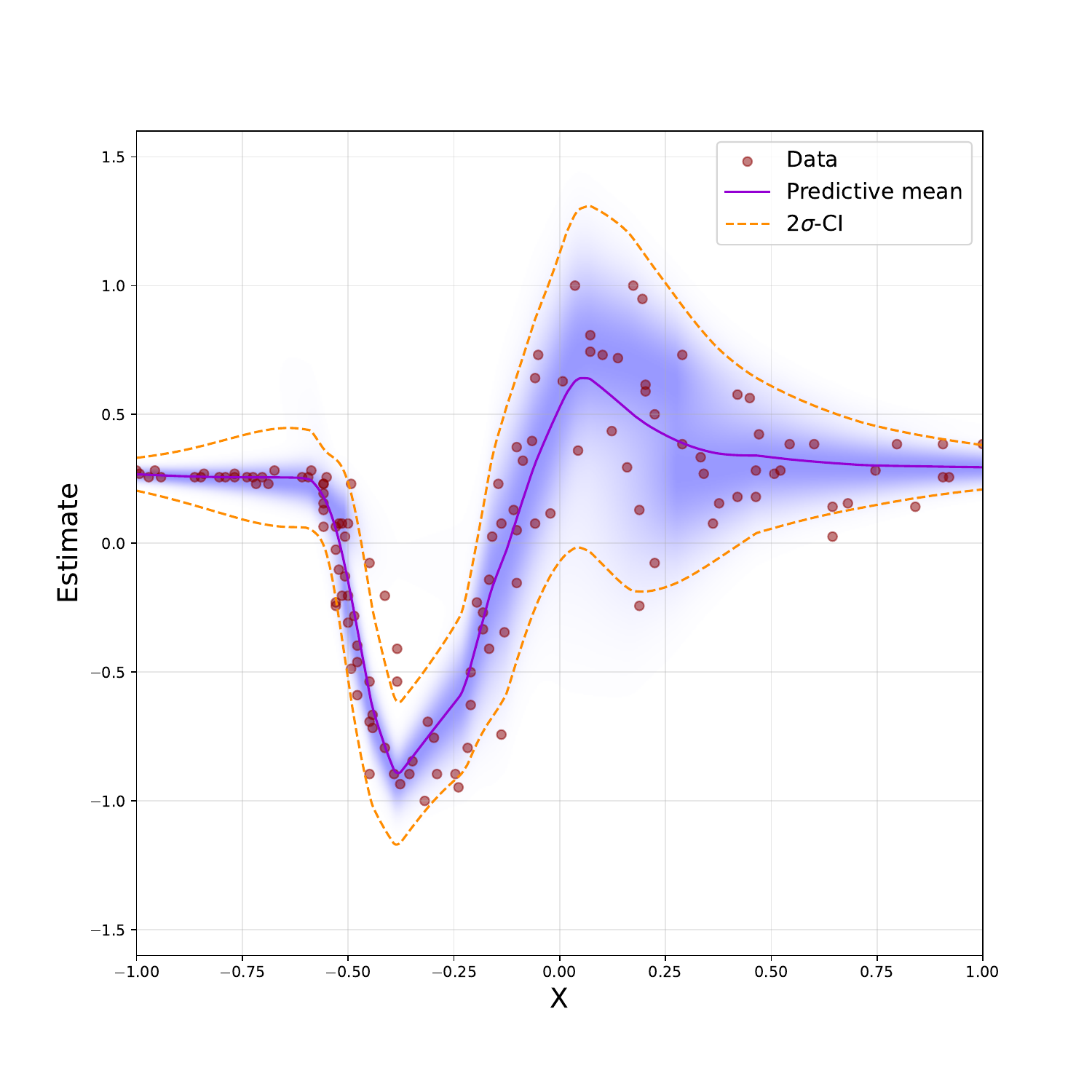}}
	\subfloat[gPoE]{\includegraphics[width=0.27\textwidth]{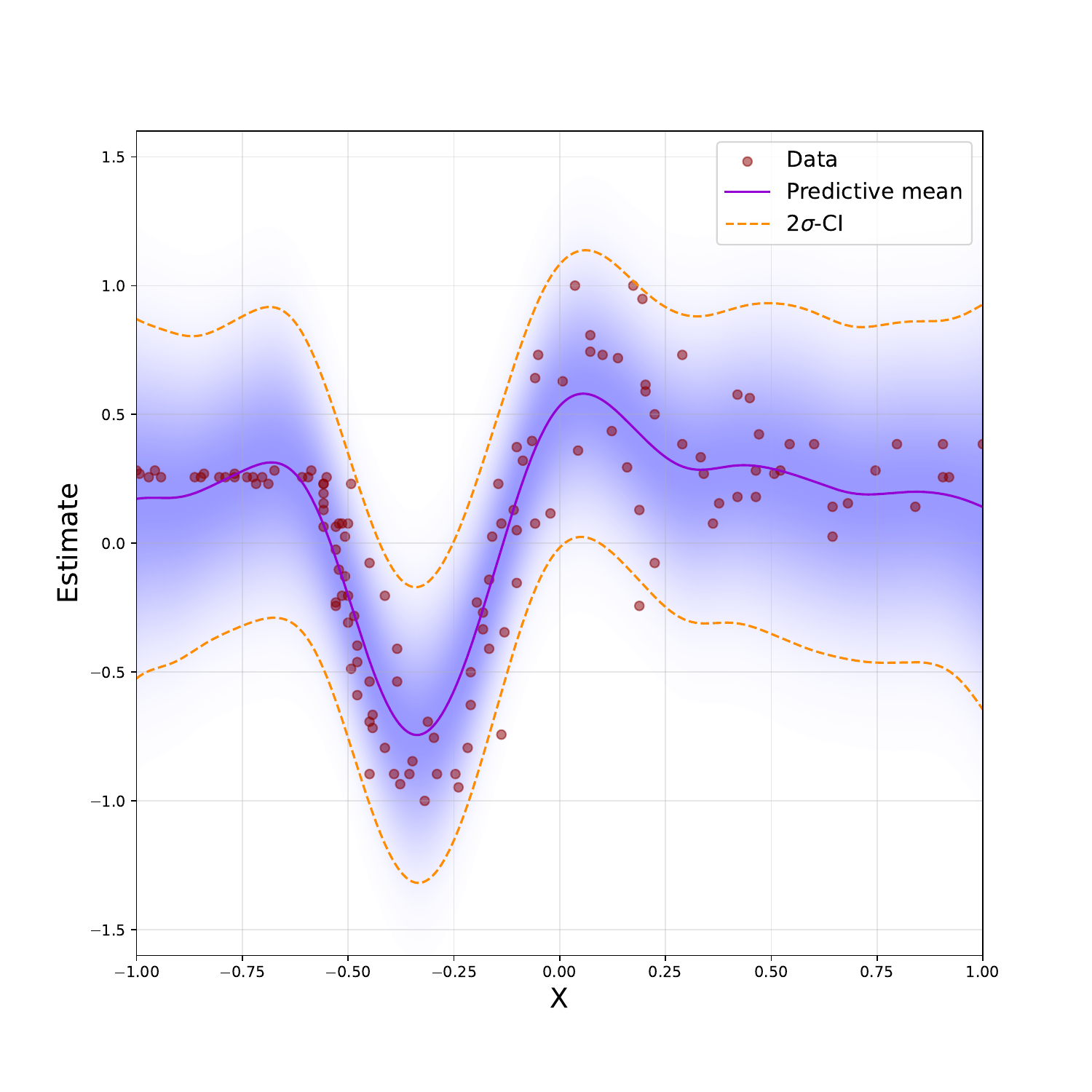}}\\
	
	\subfloat[RBCM]{\includegraphics[width=0.27\textwidth]{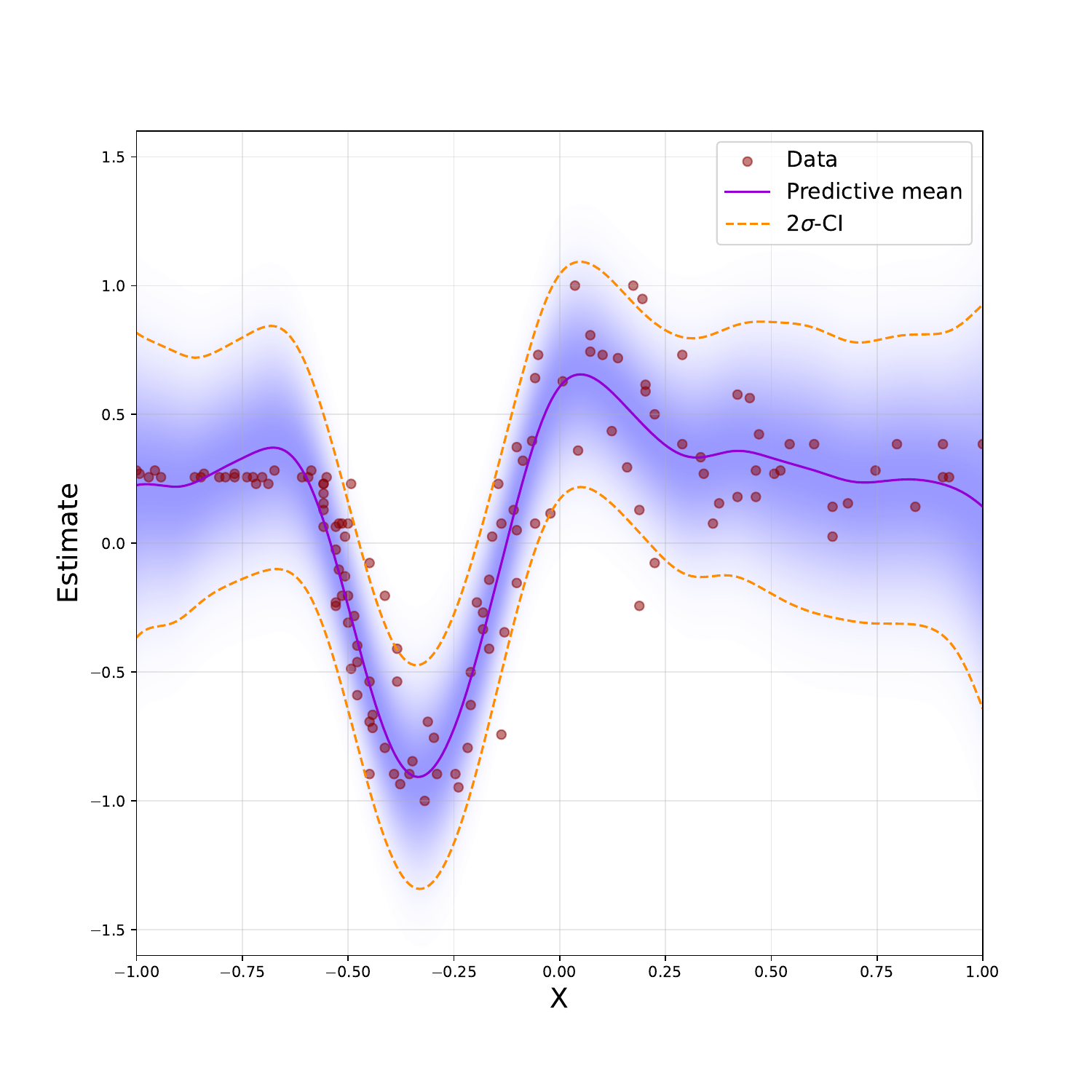}}
	\subfloat[FastGP]{\includegraphics[width=0.27\textwidth]{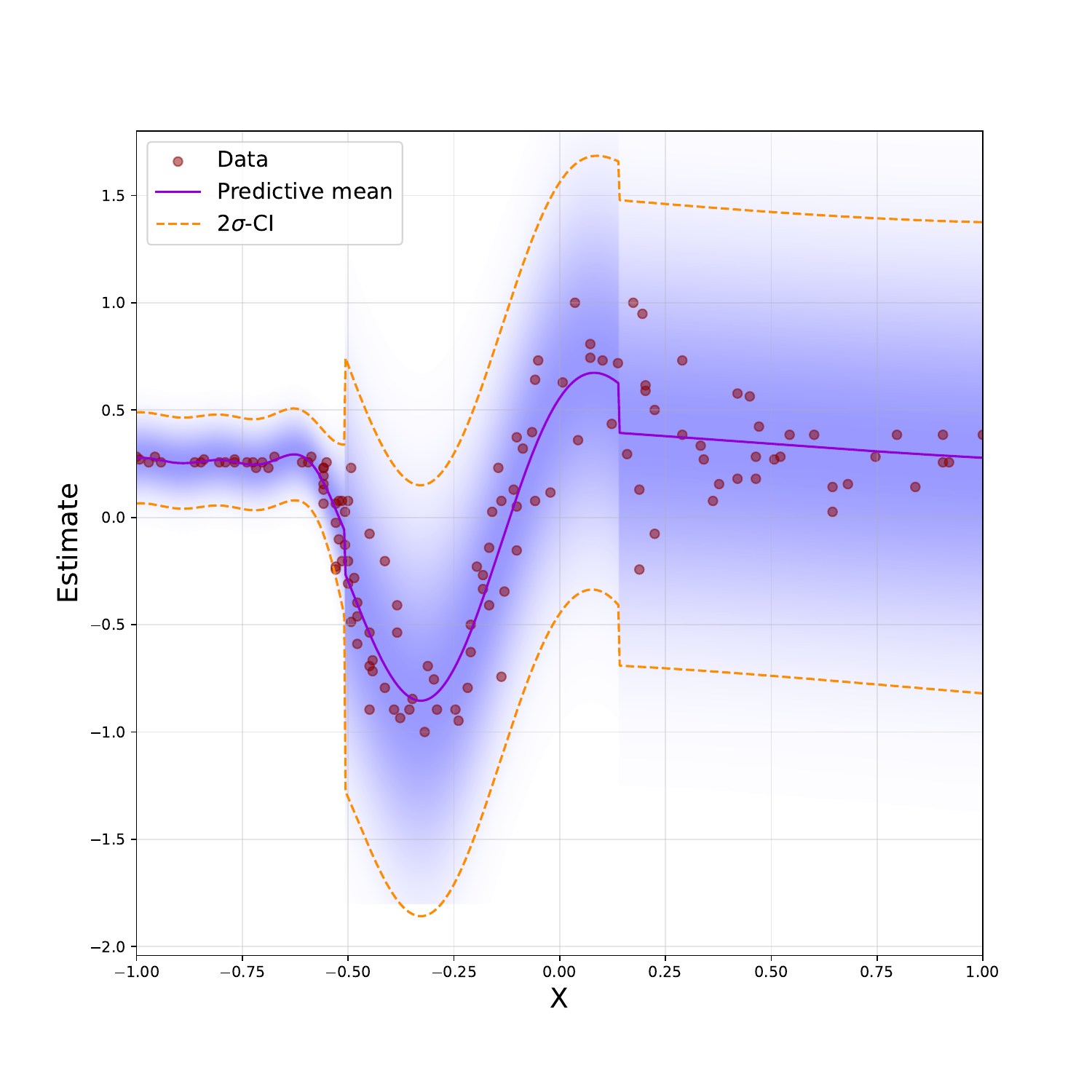}}
	\subfloat[TGP]{\includegraphics[width=0.27\textwidth]{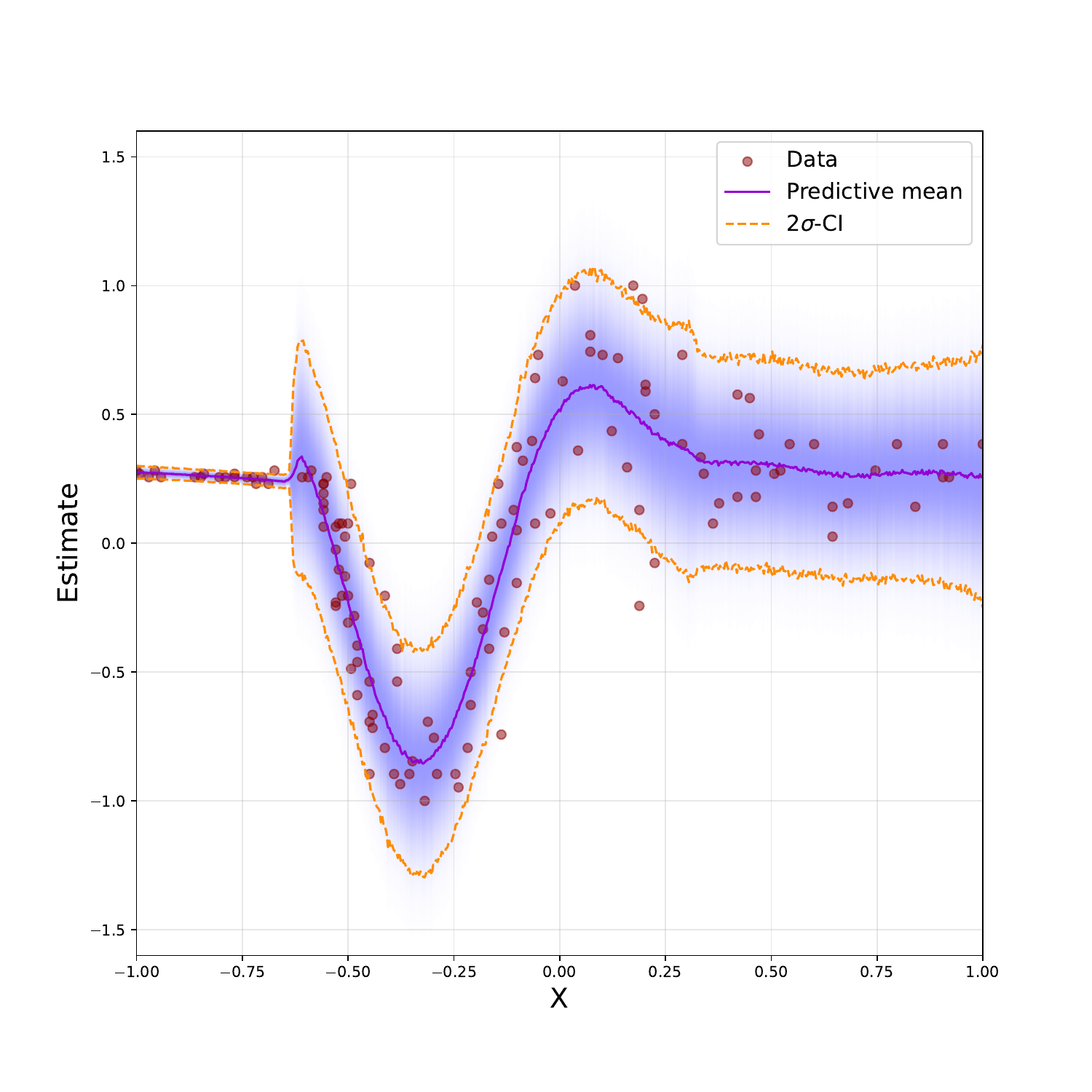}}\\
	
	\subfloat[BART]{\includegraphics[width=0.27\textwidth]{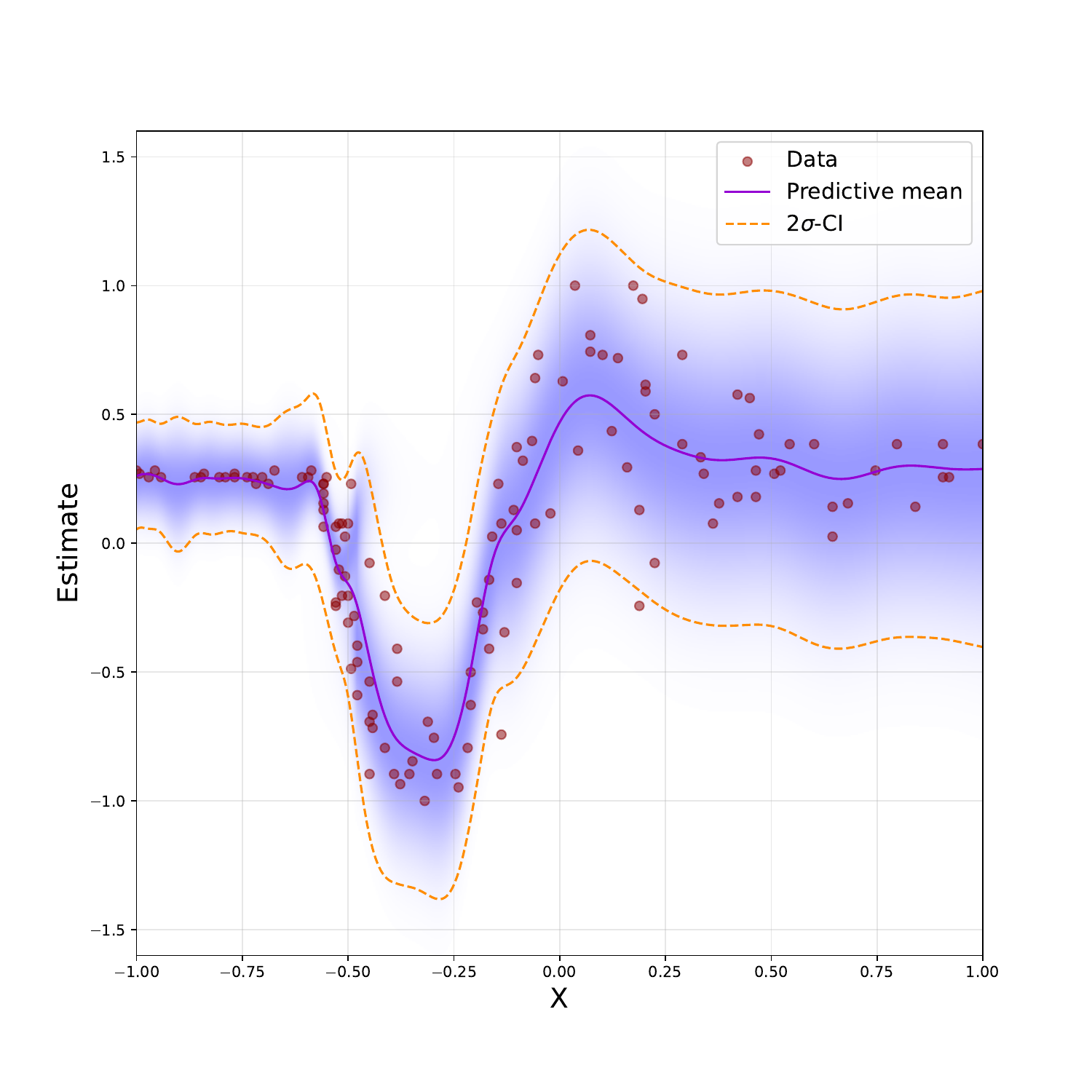}}
	\subfloat[ORTHNAT]{\includegraphics[width=0.27\textwidth]{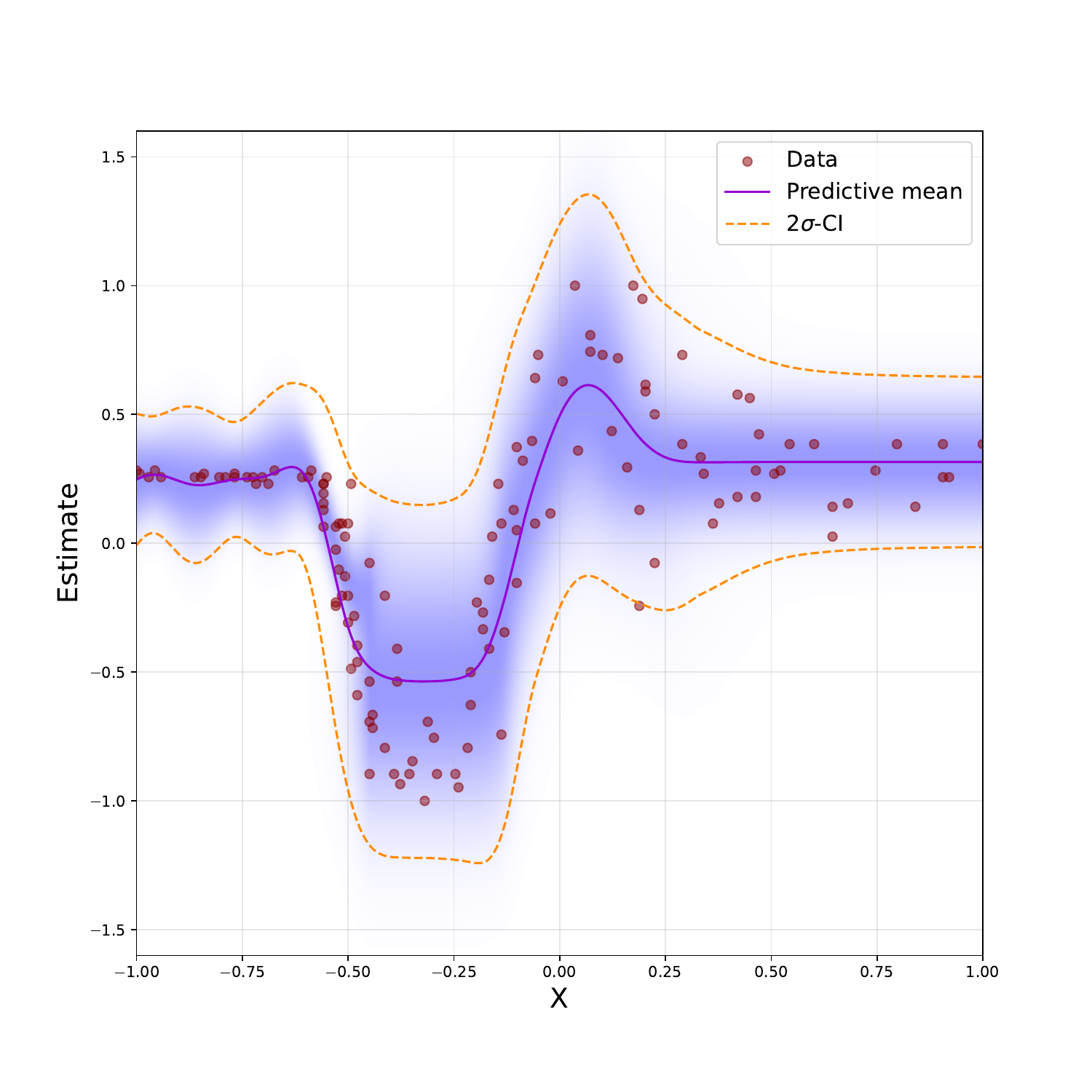}}
	\subfloat[PPGPR]{\includegraphics[width=0.27\textwidth]{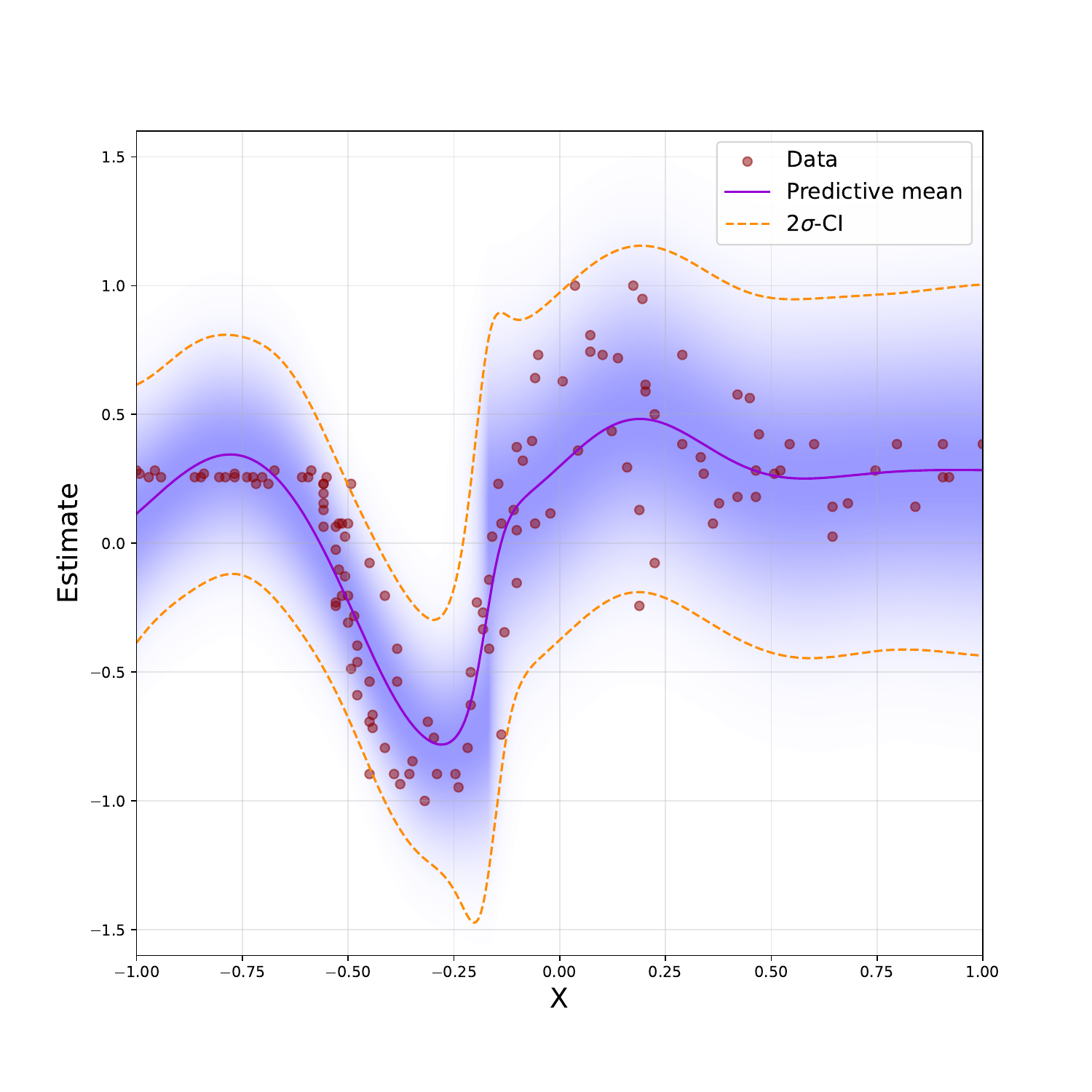}}\\
	
	\subfloat[DSPP]{\includegraphics[width=0.27\textwidth]{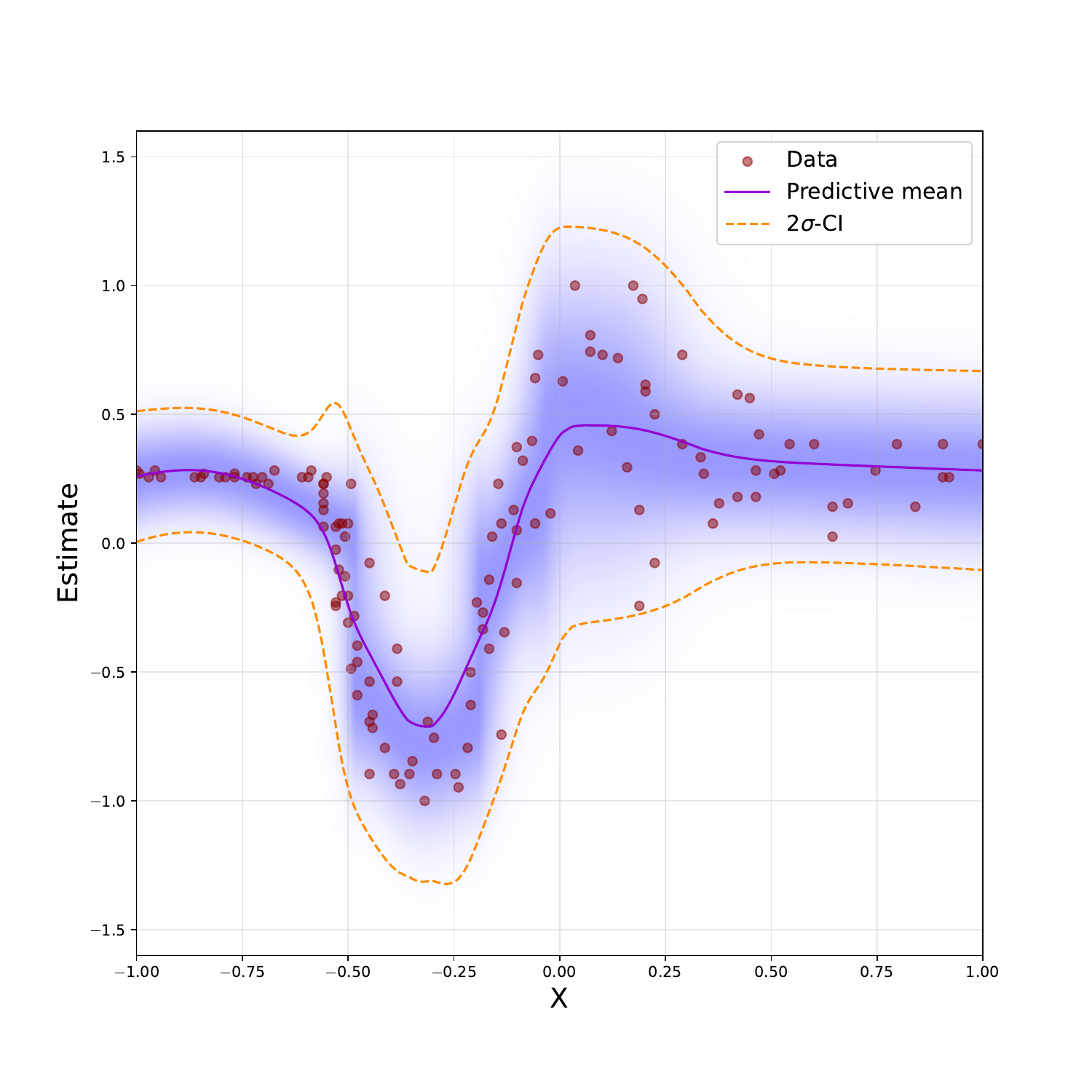}}
	\subfloat[Deep GP]{\includegraphics[width=0.27\textwidth]{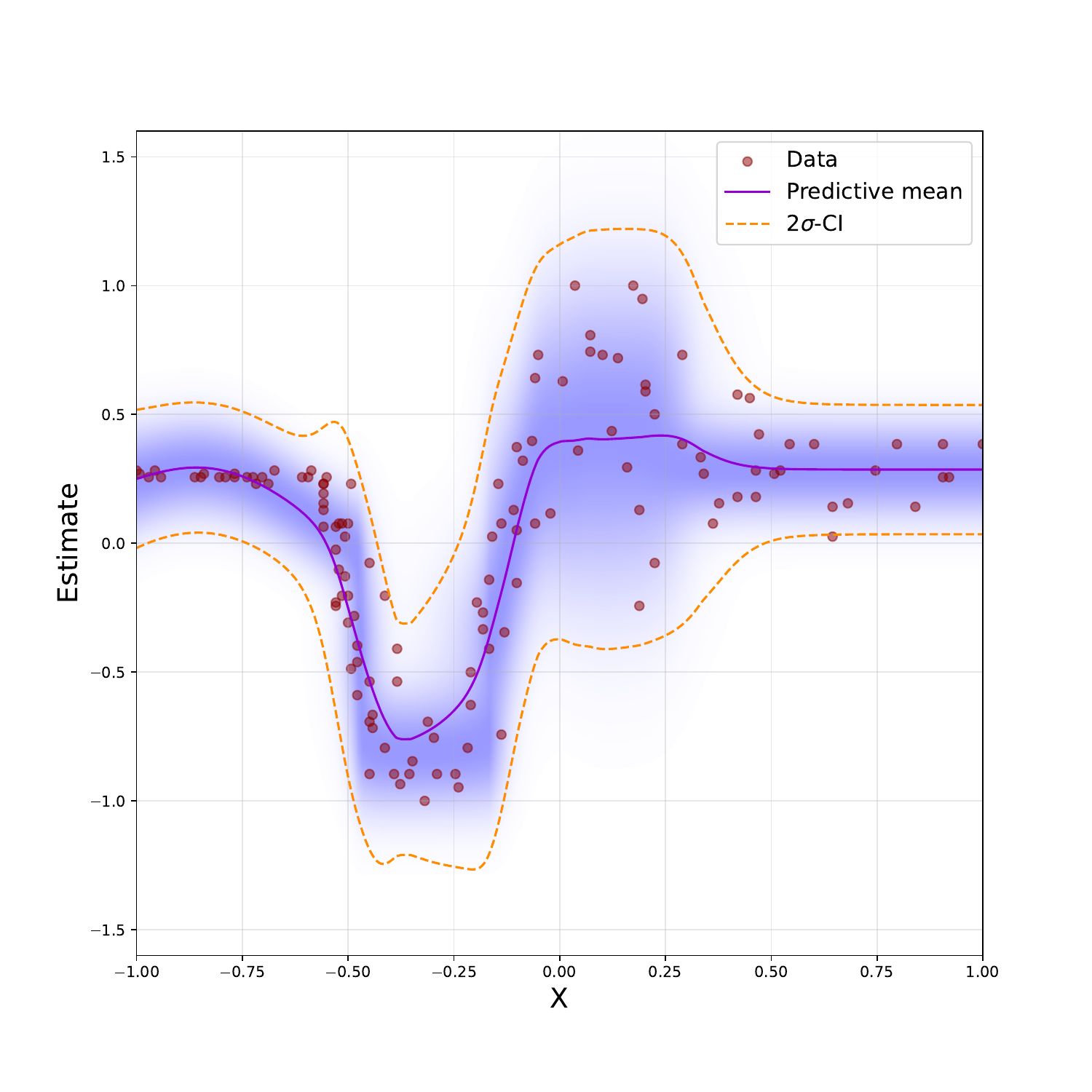}}
	\caption{Heat map of the conditional density for Motorcycle data.}\label{fig:density_plots}
\end{figure}

Lastly, we highlight that our model provides uncertainty quantification (as well as density estimation which can capture multi-modality in the predictions) for both the unknown function and predictions. \cref{fig:moto_uq} displays the soft and hard allocation predictions together with the respective 2-$\sigma$ CIs for the Motorcycle dataset. The model is able to recover the apparent non-stationary and heteroscedasticity in the data, while the other models (similar plots provided in the \cref{fig:density_plots}) tend to produce intervals that are too wide (especially on the left for $x \leq -0.6$ for PoEs) or too tight. For our model, the data (in red) is contained within the region bounded with dashed lines, suggesting that the model also provides good empirical coverage. \cref{tab:three} gives more insights into the empirical coverage (i.e. fraction of times the test points are contained within the 95\% CIs) against the average length of the CIs. It can be seen that the proposed model achieves tight intervals at the desired coverage, whereas most of the other models produce overconfident predictions or conservative intervals that are unnecessarily wide. 

\begin{figure}[htb]
	\centering	
    	\includegraphics[width=0.47\columnwidth]{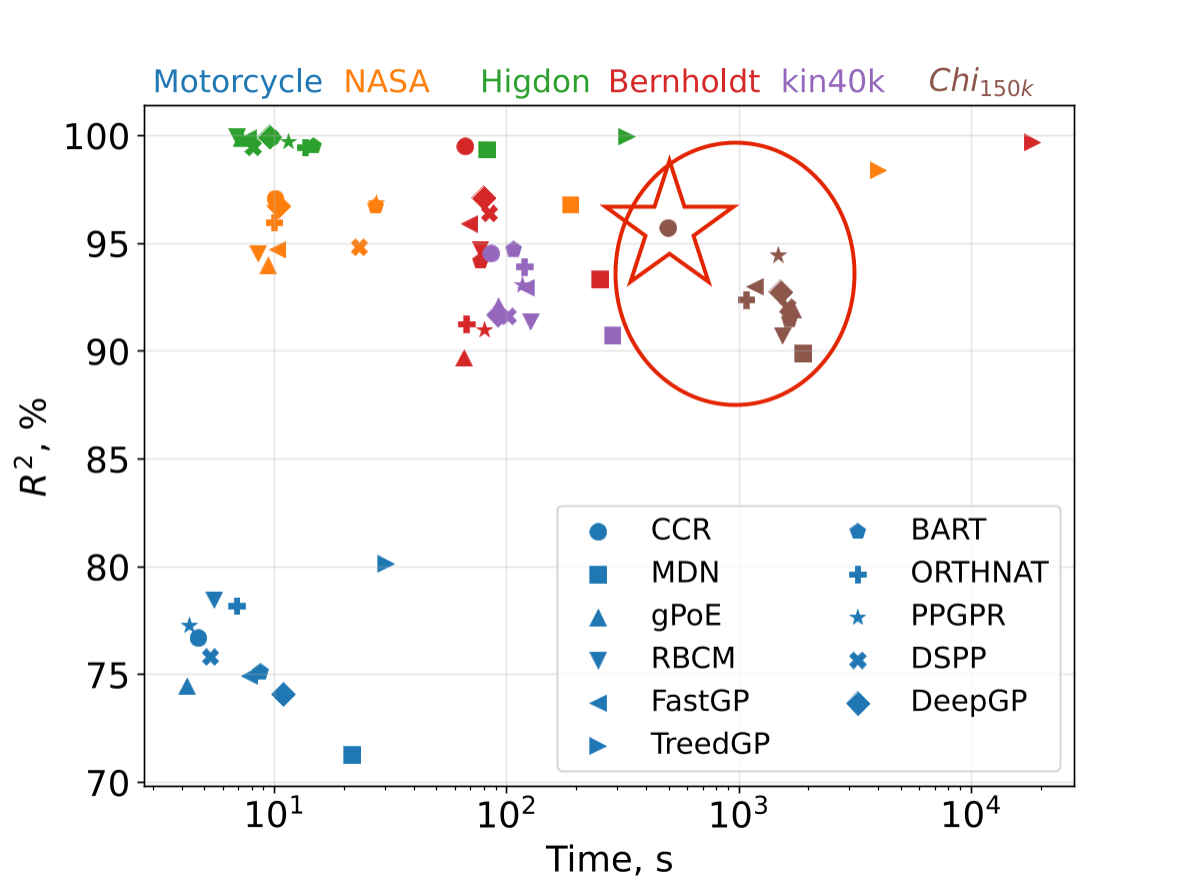}
	\includegraphics[width=0.52\columnwidth]{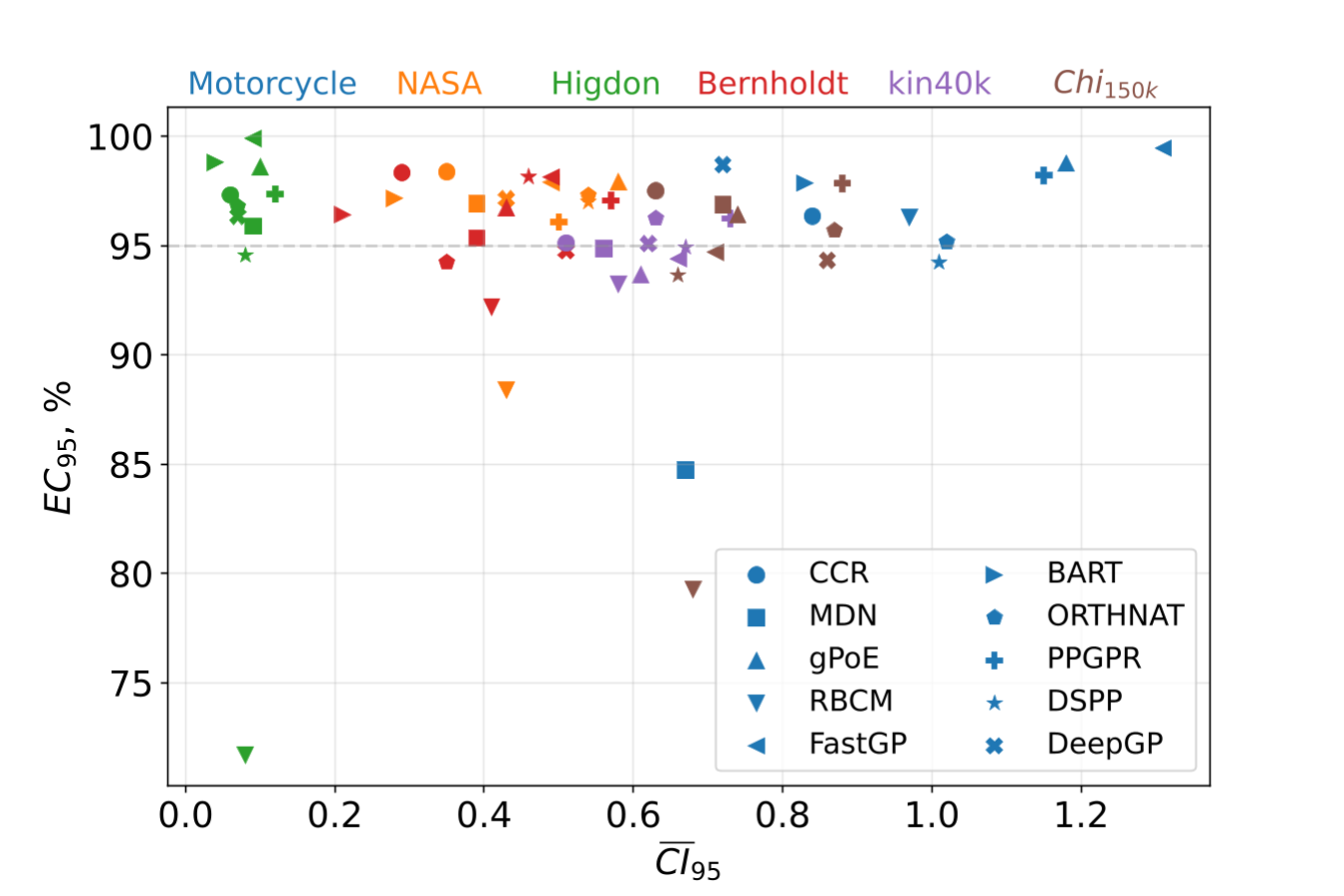}
	\caption{Left: Accuracy vs Time 
	({\em on log-scale} -- note purple/circled data).
	CCR delivers
	{\em comparable/higher accuracy, with comparable/smaller cost}. 
	Right: Empirical coverage vs Average length of 95\% CIs. CCR provides {\em judicious UQ}.}\label{fig:1}
\end{figure}

\begin{table}[htb]
\centering
\caption{Treed GP accuracy ($R^2$) on the test data, Wall-clock time, Empirical coverage ($\text{EC}_{95}$), and average length of 95\% CIs ($\bar{\text{CI}}_{95}$).}
\begin{tabular}{ | l | c | c | c | c |}
\hline
\textbf{Metric}         & $\textbf{R}^2, \%$   & \textbf{Time, s}    & $\textbf{EC}_{95}, \%$        & $\bar{\textbf{CI}}_{95}$  \\ 
\hline
\textbf{Motorcycle} 	& 80.14                     & 30.3                      & 97.86                                      & 0.83  \\
\textbf{Nasa }           	& 98.39                     & 3987.8                   & 97.16                                       & 0.28 \\
\textbf{Higdon }       	& 99.95                     & 329.5                   & 98.81                                       & 0.04 \\
\textbf{Bernholdt }   	& 99.68                     	& 18363.6                 & 96.42                                      & 0.21 \\
\textbf{kin40k }   	& -                        	& -                	       & -                                      		& - \\
$\chi_{\text{150k}}$ 	& -                             & -                           & -                                               & -  \\
\hline
\end{tabular}\label{tab:tgp}
\end{table}

\cref{tab:tgp} shows the metrics obtained for the treed GP model on all datasets but kin40k and $\chi_{\text{150k}}$ because of the computational infeasibility (the treed GP package employs MCMC inference), and hence it was separated in the standalone table. Although the treed GP model has the advantage in terms of $R^2$ and empirical coverage/interval length over CCR in several instances, it is offset by prohibitively longer execution times. 

\cref{fig:density_plots} compares heat maps of the conditional density for Motorcycle dataset. It can be seen that, unlike the other models, the proposed method successfully recovers non-stationary and heteroscedasticity in the data and provides UQ, which is closer to the one provided by the treed GP model and at a fraction of the cost. 

The results from \cref{tab:oneb,tab:two,tab:three} are presented in a more digestible format in \cref{fig:1}. \cref{fig:1} (as well as \cref{tab:oneb,tab:two,tab:three}) shows that proposed method provides the best balance between low computational cost and high accuracy, and good uncertainty quantification (nominal coverage and tight CIs) in the examples considered.

\section{Conclusion}
\label{sec:conc}

We have proposed a novel MoE, which combines powerful DNNs to flexibly determine the local regions and sparse GPs to probabilistically model the local regression functions. Through various experiments, we have demonstrated that this combination provides a flexible, robust model that is able to recover challenging behaviors such as discontinuities, non-stationarity, and non-normality and well calibrated uncertainty. In addition, we have established a novel connection between the maximization-maximization algorithm and the recently introduced CCR algorithm. This allows us to obtain a fast approximation that significantly outperforms competing methods. Moreover, in some cases, the solution can be further refined through additional MM iterations.  While we focus on the proposed deep mixture of sparse GP experts, this connection can be generally applied to other MoE architectures for fast approximation. 
Future research will explore extensions to infinite mixtures of experts, that allow for a data-driven number of clusters which can grow unboundedly with the data. 
In addition, the recent work of \citep{pmlr-v130-rossi21a} found that combining the FITC approximation of GPs with Bayesian treatment of the inducing points and hyperparameters can improve performance significantly, in particular, compared with variationally sparse GPs; in this direction, future work will also explore suitable priors for inducing variables and GP hyperparameters in MAP estimation.

\bibliography{references}
\bibliographystyle{apalike}

\appendix

\section{Number of experts}\label{app:numexperts}

\cref{tab:num_experts} contains the number of experts $L$ used for the inference for each dataset. This quantity was chosen using the BIC criterion. 

\begin{table*}[h]
\centering
\caption{Number of experts chosen with the BIC criterion.}
\begin{tabular}{ | c | c | c | c | c | c | c | c |}
\hline
\textbf{Dataset}      	& \textbf{Motorcycle} 	& \textbf{NASA}	& \textbf{Higdon}	& \textbf{Bernholdt}	& \textbf{kin40k}	& $\chi_{\text{150k}}$ \\ 
\hline
\textbf{Num of experts, $L$} 		& 6      				& 5				& 4				& 9      			& 14      			& 73 \\
\hline
\end{tabular}
\label{tab:num_experts}\end{table*}

\begin{figure}[!h]
	\centering
	\subfloat[CCR]{\includegraphics[width=0.27\textwidth]{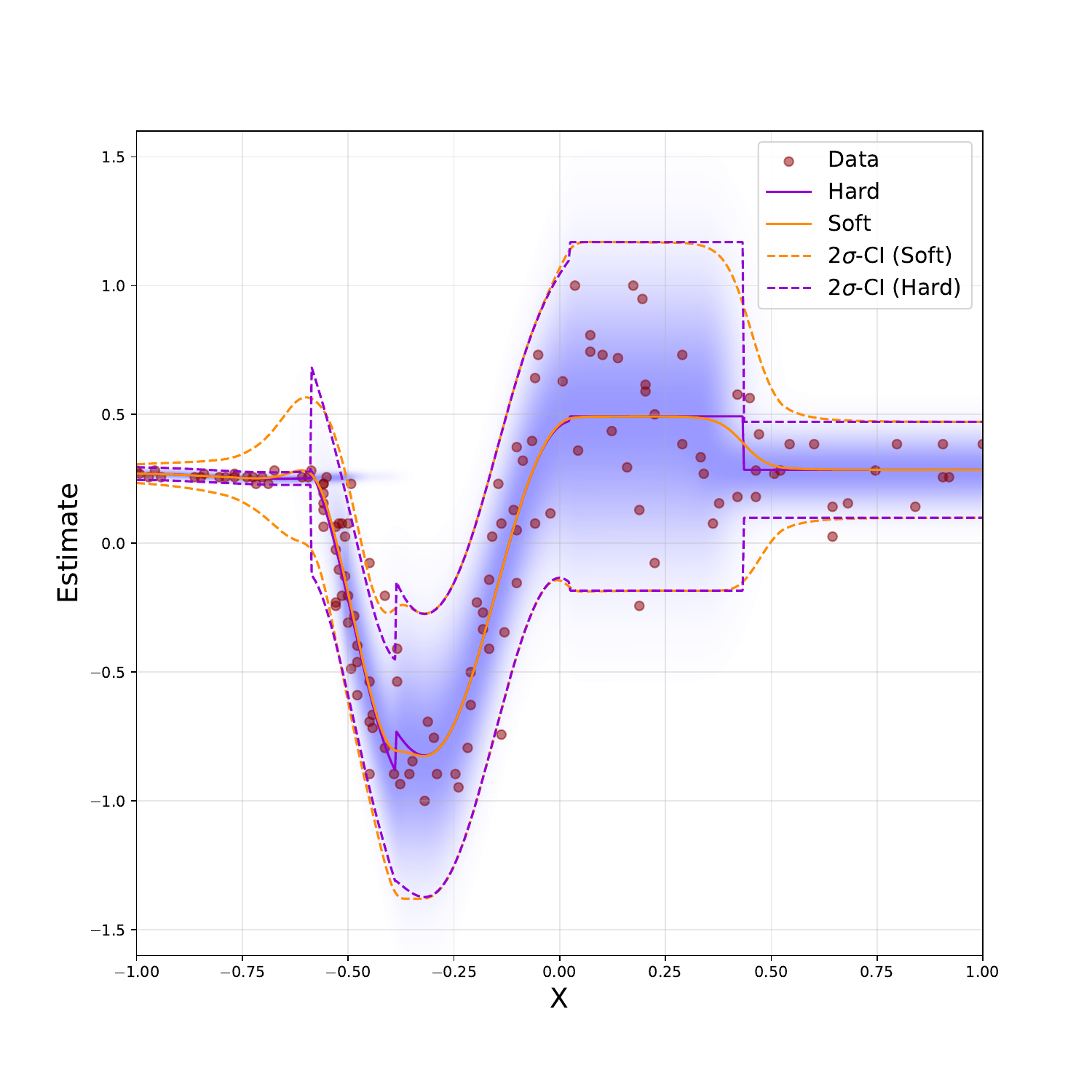}}
	\subfloat[CCR-MM]{\includegraphics[width=0.27\textwidth]{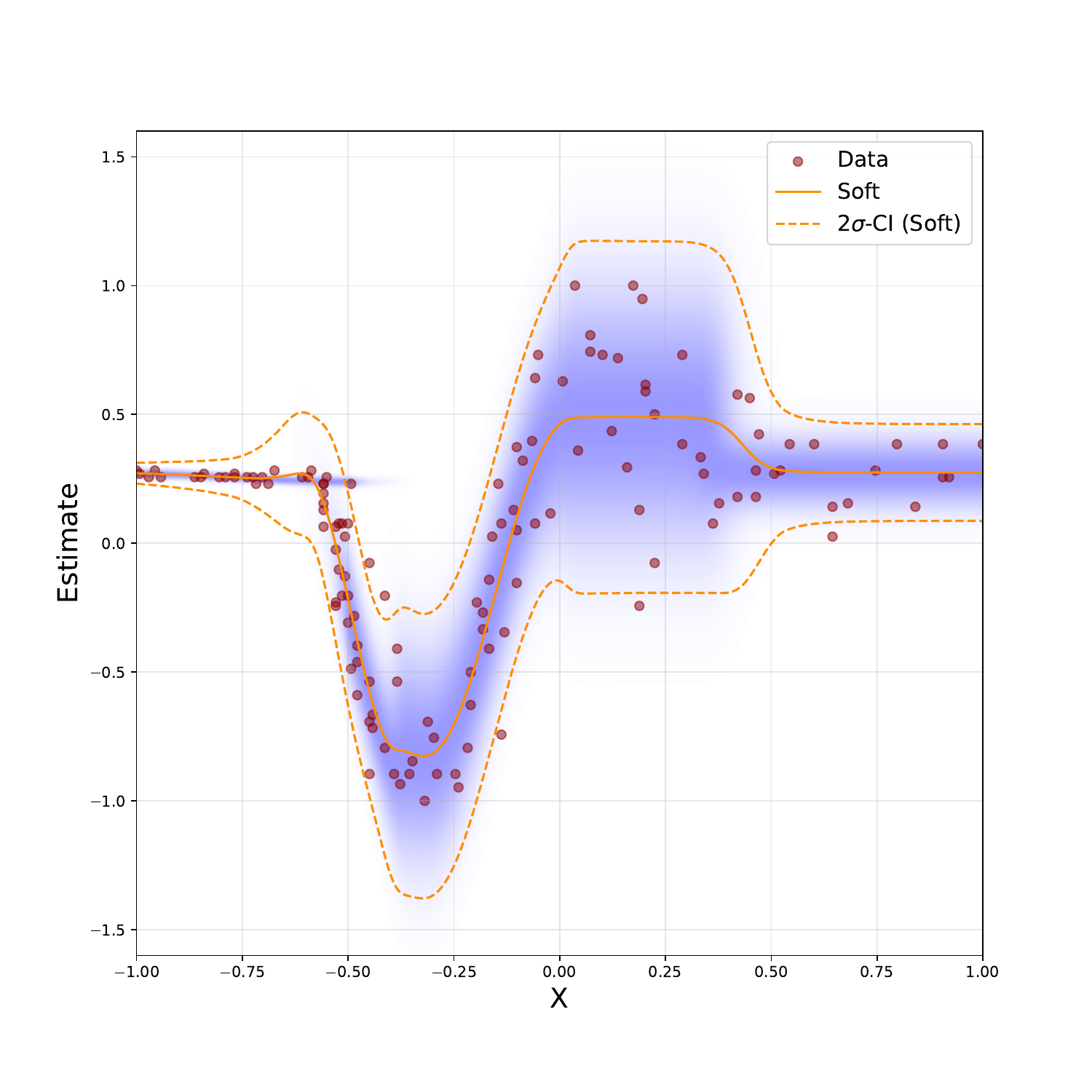}}
	\subfloat[MDN]{\includegraphics[width=0.27\textwidth]{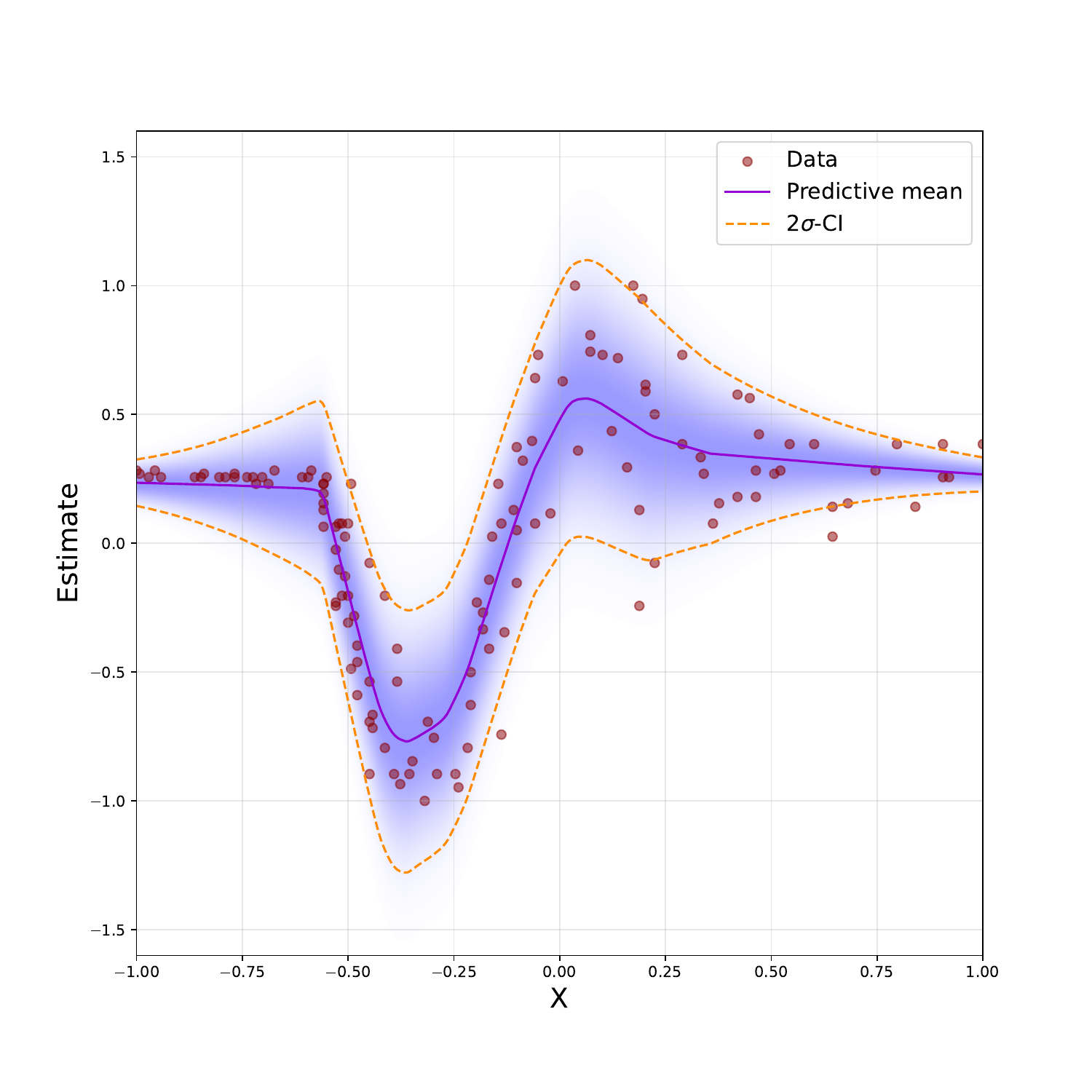}}
	\subfloat[FastGP]{\includegraphics[width=0.27\textwidth]{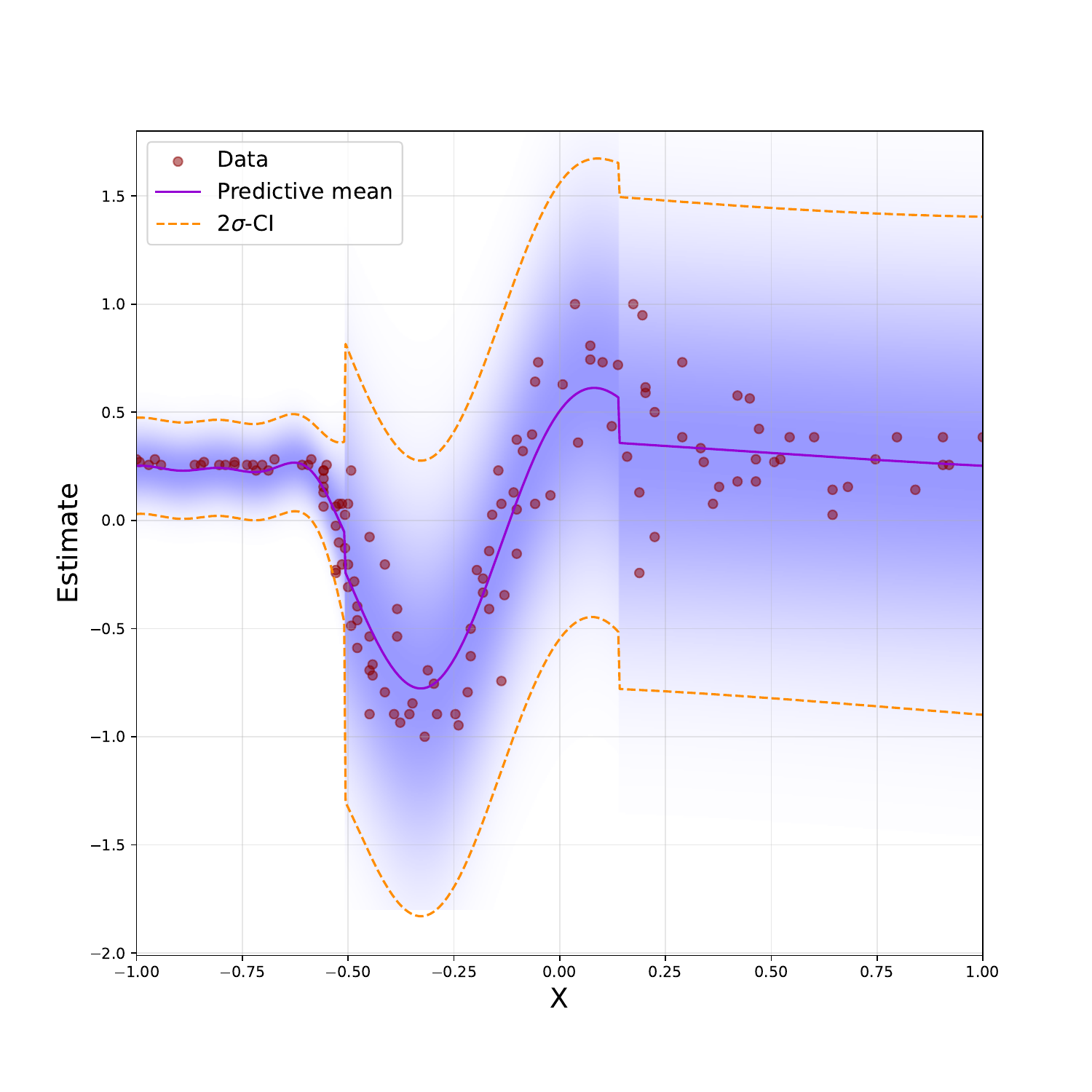}}
	\caption{Heat map of the conditional density for Motorcycle data.}\label{fig:density_plots_cv}
\end{figure}

The number of experts can also be estimated using the cross-validation. In \cref{tab:cv_experts} and \cref{fig:density_plots_cv} we present results for the Motorcycle dataset when $L$ is determined using this approach. \cref{tab:cv_experts} reports the wall-clock times taken to estimate the number of experts using both the BIC ($L_{BIC}$) and cross-validation ($L_{CV}$) approaches.

\begin{table*}[htb]
\centering
\caption{Results for the Motorcycle dataset, when the number of experts is chosen using the cross-validation.}
\begin{tabular}{ |l | c | c | c | c |}
\hline
\textbf{Model}      						& \textbf{CCR} 		& \textbf{CCR-MM}	& \textbf{MDN}		& \textbf{FastGP} \\ 
\hline
\textbf{Num of experts, $L_{CV}$} 		& 5      			& 5				& 5				& 5\\
$\textbf{R}^2, \%$						& 74.32      		& 75.17			& 70.89			& 72.43\\
\textbf{Time $L_{BIC}$, s}							& 0.54      			& 0.54			& 0.54			& 0.54\\
\textbf{Time $L_{CV}$, s}							& 17.9      			& 126.1			& 74.6			& 29.3\\
\hline
\end{tabular}
\label{tab:cv_experts}\end{table*}

\end{document}